\pdfoutput=1

\documentclass[11pt]{article}

\usepackage{acl}

\usepackage{times}
\usepackage{latexsym}

\usepackage[T1]{fontenc}

\usepackage[utf8]{inputenc}

\usepackage{microtype}

\usepackage{inconsolata}

\usepackage{algorithm}
\usepackage{algpseudocode}
\usepackage{amsmath}
\usepackage{tabularx}
\usepackage{booktabs}
\usepackage{subcaption}
\usepackage{multirow}
\usepackage{amsfonts,amssymb,bbding,pifont,makecell,bm}

\usepackage{newfloat}
\usepackage{listings}
\usepackage{graphicx}
\usepackage{xurl}
\usepackage{xparse}
\usepackage{array}
\usepackage{stfloats}

\usepackage{helvet}
\usepackage{colortbl}  
\usepackage{xcolor}
\usepackage[normalem]{ulem}
\usepackage{ragged2e}
\usepackage{CJKutf8}
\newcommand{\cellhl}{\cellcolor[HTML]{F5F5F5}}

\title{GenTranslate: Large Language Models are Generative Multilingual\\ Speech and Machine Translators}

\author{Yuchen Hu$^{1}$ \quad Chen Chen$^{1}$\quad Chao-Han Huck Yang$^{2,3}$ \\
\textbf{Ruizhe Li$^{4}$ \quad Dong Zhang$^{5}$\quad Zhehuai Chen$^{3}$\quad Eng Siong Chng$^{1}$} \\
$^1$Nanyang Technological University \quad $^2$Georgia Institute of Technology \quad $^3$NVIDIA\\$^4$University of Aberdeen \quad $^5$Fudan University \\
}

\begin{document}
\begin{CJK}{UTF8}{gbsn}

\maketitle
\begin{abstract}

Recent advances in large language models (LLMs) have stepped forward the development of multilingual speech and machine translation by its reduced representation errors and incorporated external knowledge.
However, both translation tasks typically utilize beam search decoding and top-1 hypothesis selection for inference. These techniques struggle to fully exploit the rich information in the diverse $N$-best hypotheses, making them less optimal for translation tasks that require a single, high-quality output sequence. 
In this paper, we propose a new generative paradigm for translation tasks, namely ``GenTranslate'', which builds upon LLMs to generate better results from the diverse translation versions in $N$-best list. Leveraging the rich linguistic knowledge and strong reasoning abilities of LLMs, our new paradigm can integrate the rich information in $N$-best candidates to generate a higher-quality translation result. 
Furthermore, to support LLM finetuning, we build and release a HypoTranslate dataset that contains over 592K hypotheses-translation pairs in 11 languages.
Experiments on various speech and machine translation benchmarks (\emph{e.g.}, FLEURS, CoVoST-2, WMT) demonstrate that our GenTranslate significantly outperforms the state-of-the-art model\footnote{This work is open sourced at: \url{https://github.com/YUCHEN005/GenTranslate}}.

\end{abstract}


\section{Introduction}
Recent advances in large language models (LLMs) have attracted a surge of research interest due to their strong abilities in logical reasoning and language generation \citep{chatgpt,gpt4,touvron2023llama,touvron2023llama2}. These models have achieved surprisingly wide-ranging success across various natural language processing (NLP) tasks~\citep{brown2020language,wang2022language,wei2022emergent,wei2022chain,ouyang2022training}.

\begin{figure}[t]
\begin{center}
\includegraphics[width=0.97\columnwidth]{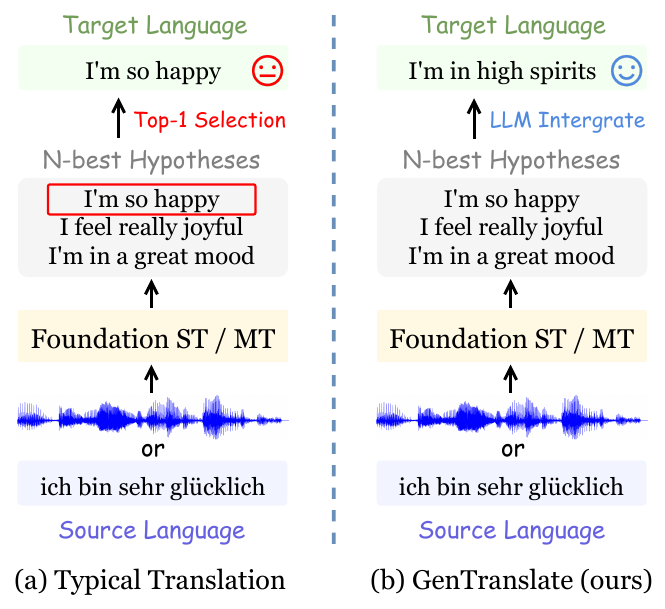}
\end{center}
\vspace{-0.35cm}
\caption{Illustration of (a) Typical seq2seq translation with beam search decoding and top-1 hypothesis selection, (b) our ``GenTranslate'' with LLM integration.}
\vspace{-0.5cm}
\label{fig1}
\end{figure}

In the realm of NLP, the translation tasks, which encompasses speech and machine translation (ST \& MT), hold significant practical importance for global communication.
Similar to other NLP tasks, translation tasks also gain a notable progress thanks to the recent advancement of LLMs~\citep{zhang2023prompting,lyu2023new}.
In the domain of speech translation, Whisper~\citep{radford2023robust} demonstrates superior performance by collecting 680K-hour data for web-scale model training.
AudioPaLM2~\citep{rubenstein2023audiopalm} integrates both text- and speech-based language models into a unified architecture to process and generate text and speech, thereby augmenting speech translation performance to a great extent.
On the other hand, LLMs also show remarkable ability in machine translation.
NLLB~\citep{costa2022no} is the first to extend LLMs' linguistic capability to over 200 languages.
BigTranslate~\citep{yang2023bigtrans} is finetuned on LLaMA~\citep{touvron2023llama} with multilingual instruction tuning, which achieves comparable performance to ChatGPT~\cite{chatgpt} and Google Translate.
Most recent work proposes SeamlessM4T~\citep{barrault2023seamlessm4t}, a foundational multilingual and multitask model that can translate across speech and text, which achieves the state-of-the-art on both ST and MT tasks on various public datasets.

\begin{figure}[t]
\begin{center}
\includegraphics[width=0.97\columnwidth]{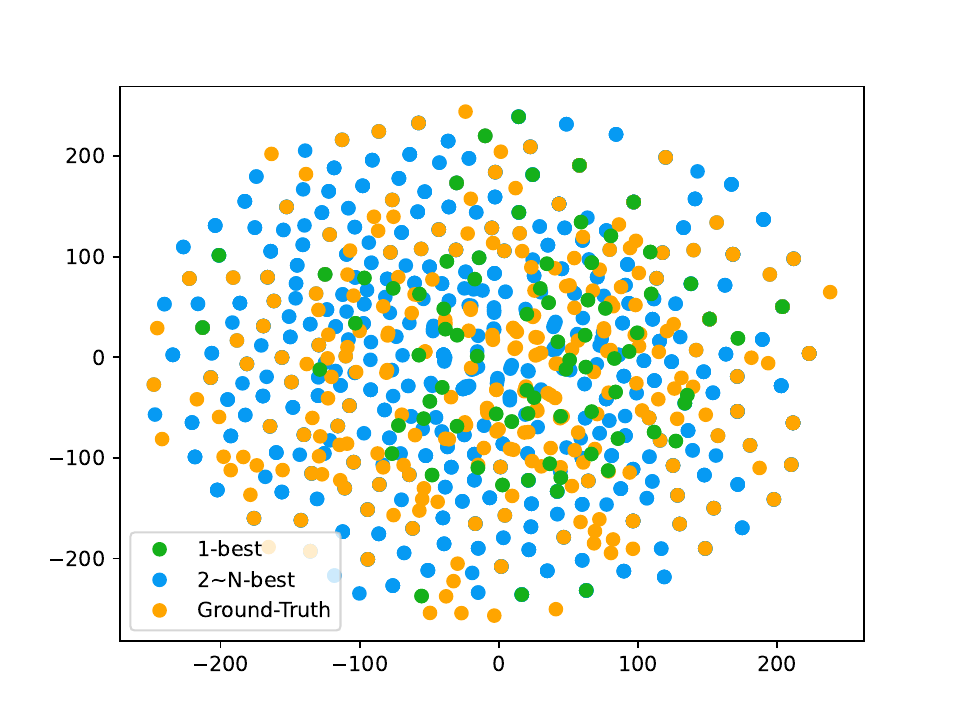}
\end{center}
\vspace{-0.2cm}
\caption{t-SNE visualization of the n-gram tokens (n=1,2,3) in ST 1-best hypothesis ({\color{green}green}), 2 to $N$-best hypotheses ({\color{blue}blue}), and the ground-truth translation ({\color{orange}orange}), where the text embeddings are extracted using SBERT~\citep{reimers2019sentence}.
It indicates that the 2 to $N$-best hypotheses contain richer information than 1-best for generating ground-truth translation.}
\vspace{-0.3cm}
\label{fig2}
\end{figure}

Despite the superior performance, most existing translation models employ the typical beam search algorithm for inference and select the top-1 hypothesis as final output (see Fig.~\ref{fig1} (a)), following that in automatic speech recognition (ASR)~\citep{tsunoo2021streaming}.
However, this strategy discards the 2 to $N$-best hypotheses that could be advantageous to the generation of ground-truth translation.
As illustrated in Fig.~\ref{fig2}, the discarded 2 to $N$-best hypotheses contain abundant semantic information that is the key to composite the ground-truth utterance, while the 1-best hypothesis lacks this part of information.
As a result, the typical top-1 hypothesis selection is sub-optimal to the translation tasks that require a single informative and high-quality output sequence~\citep{li2022diffusion,xiao2022bitiimt}.

Inspired by the recent works on LLMs-enhanced ASR~\citep{ma2023can,chen2023hp,yang2023generative,radhakrishnan2023whispering}, we propose a new generative paradigm for translation tasks, namely GenTranslate (see Fig.~\ref{fig1} (b)).
Leveraging the rich linguistic knowledge and strong reasoning ability of LLMs, our paradigm integrates the diverse translation versions in the $N$-best list from foundation model to generate a higher-quality translation result.
Furthermore, in order to support LLM finetuning, we also build and release a HypoTranslate dataset that contains over 592K pairs of $N$-best hypotheses and ground-truth translation in 11 languages.
Experimental evidence on various ST and MT benchmarks (\emph{e.g.}, FLEURS, CoVoST-2, WMT) demonstrate that our proposed GenTranslate significantly outperforms the state-of-the-art model with efficient LLM finetuning.

Our contributions are summarized as follows:
\begin{itemize}
    \item We propose GenTranslate, a new generative paradigm for translation tasks that leverages LLMs to generate higher-quality translation results from the diverse $N$-best hypotheses decoded from foundation translation model.
    \item We release a HypoTranslate dataset to support LLM finetuning, which contains over 592K pairs of $N$-best hypotheses and ground-truth translation in 11 languages.
    \item Experiments on various ST and MT benchmarks show that our GenTranslate significantly outperforms the state-of-the-art model.
\end{itemize}

\section{Related Work}
\label{sec:related_work}

\subsection{Large Language Models}
\label{ssec:llm}
There is recently a surge of research interests in Transformer-based large language models, such as ChatGPT~\citep{chatgpt}, GPT-4~\citep{gpt4} and LLaMA~\citep{touvron2023llama,touvron2023llama2}.
Benefiting from the giant model size and oceans of training data, LLMs can  understand better the linguistic structures and semantic meanings behind raw text, which thus shows remarkable performance on a wide range of natural language processing (NLP) tasks~\citep{brown2020language,wei2022emergent,ouyang2022training}.
Thereafter, with techniques like in-context learning~\citep{xie2021explanation} and efficient finetuning~\citep{hu2021lora, yang2021voice2series}, LLMs further show powerful ability on downstream generative and reasoning tasks~\citep{lampinen2022can,yang2023generative,hu2023llm,zhang2023llama}.
Our proposed GenTranslate is exactly inspired by the promising generative ability of LLMs.

\subsection{Speech and Machine Translation}
\label{ssec:speech_machine_trans}
The advancement of LLMs has notably enhanced the capabilities of translation tasks.
In the domain of speech translation~\cite{liu2021ustc}, Whisper~\cite{radford2023robust} demonstrates commendable effectiveness, leveraging extensive web-scale data. AudioPaLM2~\cite{rubenstein2023audiopalm} integrates text- and speech-based language models, thereby augmenting the speech translation performance.
In the context of machine translation, NLLB~\cite{costa2022no}, a model fine-tuned on LLMs, extends its linguistic range to over 200 languages. Additionally, BigTranslate~\cite{yang2023bigtrans} utilizes instruction tuning to enhance the translation capabilities of LLMs.
The most recent innovation, SeamlessM4T~\citep{barrault2023seamlessm4t}, represents a  highly-unified model capable of fluid translation between speech and text, setting new benchmarks in both ST and MT tasks. However, it is noteworthy that the majority of these methodologies rely on beam search decoding~\cite{yang2021multi, hu2023improving} and top-1 hypothesis selection for inference. How to leverage $N$-best hypotheses to deliver better translation result remains to be an open question.

\subsection{LLMs-Enhanced ASR}
\label{ssec:llm_enhanced_asr}
Recent works investigate LLMs to enhance the ASR output by error correction~\citep{ma2023n,chen2023hp,chen2023generative,chen2024its,hu2024large}, which serves as a post-processing technique to improve the recognition result~\citep{leng2021fastcorrect}.
In particular, they leverage LLM finetuning~\citep{zhang2023llama} and in-context learning~\citep{wang2023can} to correct the wrongly recognized tokens in hypotheses by second-pass reasoning, which achieves promising improvement.
Inspired by them, in this work we leverage LLMs to integrate the diverse translation versions in $N$-best list to generate a informative and higher-quality translation result.

\begin{figure*}[t]
\begin{center}
\includegraphics[width=1.0\textwidth]{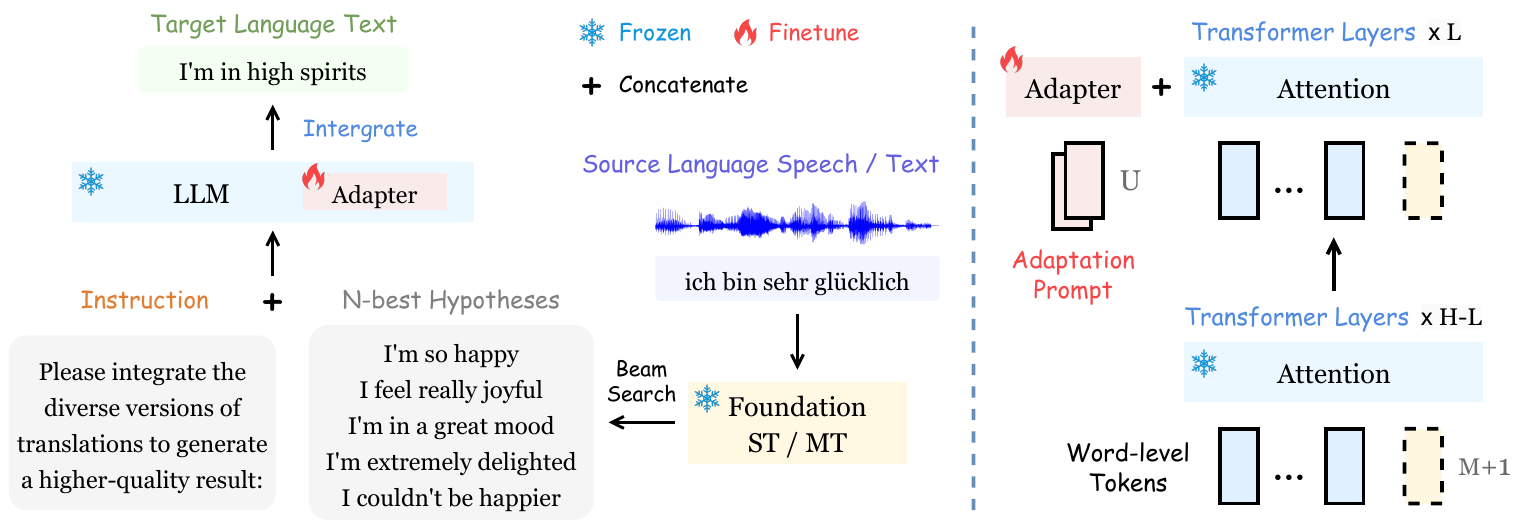}
\end{center}
\vspace{-0.2cm}
\caption{\textbf{Left:} Overview of the GenTranslate paradigm (\emph{e.g.}, De$\rightarrow$En). \textbf{Right:} Details of efficient LLM finetuning.}
\vspace{-0.4cm}
\label{fig3}
\end{figure*}

\section{Methodology}
\label{sec:method}
In this section, we introduce the proposed method.
First, we describe the latest foundational translation model, SeamlessM4T, which we employ for beam search decoding and hypotheses generation (\S\ref{ssec:seamlessm4t}).
Then, we introduce our LLMs-based GenTranslate paradigm by $N$-best hypotheses integration (\S\ref{ssec:gentrans}).
Finally, we present the details of our released HypoTranslate dataset for GenTranslate training (\S\ref{ssec:hypotrans_dataset}).

\subsection{Foundational Translation Model: SeamlessM4T}
\label{ssec:seamlessm4t}

Recent work~\citep{barrault2023seamlessm4t,barrault2023seamlessm4tv2} proposes SeamlessM4T\footnote{\url{https://github.com/facebookresearch/seamless_communication}} (Massively Multilingual \& Multimodal Machine Translation), a single Transformer-based~\citep{vaswani2017attention} model that supports speech-to-speech translation, speech-to-text translation, text-to-speech translation, text-to-text translation, and automatic speech recognition for up to 100 languages.
During development process, it is firstly pre-trained on 1 million hours of speech data by self-supervised learning, and it is then finetuned on a 406K-hour multimodal corpus of automatically aligned speech translations named SeamlessAlign.
Experiments show that SeamlessM4T yields superior performance on all of the five supported tasks.
In particular, it has achieved the state-of-the-art on both ST and MT tasks in terms of BLEU score on various public benchmarks.

Considering its effectiveness, generality and popularity, we employ SeamlessM4T as the foundation model for both speech and machine translation in our system, as depicted in the left part of Fig.~\ref{fig3}.
Given an input speech $S^\text{src}$ or text $T^\text{src}$ in source language (\emph{e.g.}, German), SeamlessM4T translates it into target language (\emph{e.g.}, English) text by beam search decoding, which generates $N$-best hypotheses list $\mathcal{T}_N^\text{tgt} = \{T_1^\text{tgt},T_2^\text{tgt},\cdots,T_N^\text{tgt}\}$.

\subsection{GenTranslate}
\label{ssec:gentrans}

\subsubsection{Overall Framework}
\label{sssec:overall}
To solve the information loss in typical top-1 hypothesis selection, we leverage LLMs to generate a final translation result based on the decoded $N$-best hypotheses.
Since each candidate in $N$-best list represents one unique version of translation for source language input, our GenTranslate can integrate their rich information to generate a higher-quality translation result, thanks to the strong linguistic and reasoning ability of LLMs.
This new generative paradigm can be formulated as:
\begin{equation}
\label{eq1}
\begin{aligned}
    T^\text{tgt} &= \mathcal{M}_\text{GT} (\mathcal{T}_N^\text{tgt}, \mathcal{I}),
\end{aligned}
\end{equation}
where $\mathcal{I}$ is a proper instruction for LLM prompting.
The goal of GenTranslate is to learn a mapping $\mathcal{M}_\text{GT}$ from $N$-best hypotheses to the true translation.
Following typical sequence-to-sequence learning strategy, we employ the ground-truth translation $T^\text{tgt*}$ as supervision signal and optimize the LLM to learn $\mathcal{M}_\text{GT}$ in an auto-regressive manner.
The cross-entropy-based training loss is defined as:
\begin{equation}
\label{eq2}
\begin{aligned}
    \mathcal{L}_\text{GT} &= \sum_{l=1}^L -\log \mathbb{P}_{\theta} (t_l^\text{tgt*} | t_{l-1}^\text{tgt*}, \cdots, t_1^\text{tgt*}; \mathcal{T}_N^\text{tgt}, \mathcal{I}),
\end{aligned}
\end{equation}
where $t_l^\text{tgt*}$ is the $l$-th token of $T^\text{tgt*}$, $L$ denotes the sequence length, and $\theta$ denotes the learnable parameters in LLM (\emph{i.e.}, adapter).

\subsubsection{Efficient LLM Finetuning}
\label{sssec:llama_adapter}
Considering the giant scale of LLMs, we adopt the popular efficient finetuning strategy, LLaMA-Adapter~\citep{zhang2023llama}, which is comparable to LoRA tuning (\S\ref{sssec:adapter_vs_lora}).
As shown in Fig.~\ref{fig3} (right), it inserts a set of learnable adaptation prompts into the top-$L$ of total $H$ Transformer layers in a pre-trained LLM to learn high-level semantics.
Denote the prompt for $l$-th layer as $\mathcal{P}_l\in\mathbb{R}^{U\times D}$, where $U$ is prompt length and $D$ is embedding size.

Assume we gain $M$ tokens including instruction and already generated response, \emph{i.e.}, $T_l\in\mathbb{R}^{M\times D}$, now we aim to predict the $(M+1)$-th token as response.
The learnable adaptation prompt is concatenated with $T_l$ as prefix, \emph{i.e.}, $[\mathcal{P}_l;T_l]\in\mathbb{R}^{(U+M)\times D}$, which provides learned instruction knowledge to guide the subsequent response generation.

Furthermore, considering the prompt $\mathcal{P}_l$ is randomly initialized and thus could disturb the LLM tuning at early training stage, a zero-initialized attention mechanism is devised to mitigate such disturbance.
Denote the current $M$-th token as $T_l^{(M)}\in\mathbb{R}^{1\times D}$, in attention there are three projection layers to generate query, key and value:
\begin{equation}
\label{eq3}
\begin{aligned}
    Q_l &= \mathrm{Linear}_q(T_l^{(M)}),\\
    K_l &= \mathrm{Linear}_k([\mathcal{P}_l;T_l]),\\
    V_l &= \mathrm{Linear}_v([\mathcal{P}_l;T_l]),
\end{aligned}
\end{equation}
Then the attention score is calculated as $A_l=Q_l\cdot K_l/\sqrt{D}\in\mathbb{R}^{1\times(U+M)}$, which captures the correlation between current token and the history tokens as well as prompts to predict the next token.
Therefore, it can be split into two parts accordingly:
\begin{equation}
\label{eq4}
    A_l = [A_l^\mathcal{P}; A_l^T]^T,
\end{equation}
where $A_l^\mathcal{P}\in\mathbb{R}^{U\times 1}$ is the attention score of $U$ adaptation prompts and $A_l^T\in\mathbb{R}^{M\times 1}$ is that of $M$ history tokens.
Since the adaptation prompts are randomly initialized, their attention scores may cast disturbance on next-token prediction at early training stage.
To this end, a learnable gating factor $g_l$ with zero initialization is introduced to adaptively control the weight of prompt in attention:
\begin{equation}
\label{eq5}
    A_l^g = [g_l\cdot \mathrm{softmax}(A_l^\mathcal{P});\hspace{0.1cm}\mathrm{softmax}(A_l^T)]^T,
\end{equation}
Finally, the attention output of $l$-th Transformer layer is obtained with a linear projection:
\begin{equation}
\label{eq6}
    O_{l}^{(M)} = \mathrm{Linear}_o(A_l^g\cdot V_l) \in\mathbb{R}^{1\times D},
\end{equation}
It is then employed to predict the next token $T_{l}^{(M+1)}$ as response.
The zero-initialization mechanism yields an effective trade-off between the pre-trained knowledge of LLM and the learned instructional knowledge through adaptation prompt.

\subsection{HypoTranslate Dataset}
\label{ssec:hypotrans_dataset}
In order to support the LLM finetuning for GenTranslate, we release a HypoTranslate dataset that contains over 592K pairs of $N$-best hypotheses and ground-truth translation in 11 languages.
In particular, we use the state-of-the-art SeamlessM4T-Large as foundation translation model to decode $N$-best hypotheses from input speech by beam search algorithm, where the beam size $N$ is set to 5.
Specifically, for ST task we investigate two popular pipelines in literature, \emph{i.e.}, end-to-end ST and cascaded ASR+MT.
Thanks to the universal ability of SeamlessM4T on ST, ASR and MT tasks, we only need one model to build above two pipelines.

To build HypoTranslate dataset, we select several public ST and MT corpora in both X$\rightarrow$En and En$\rightarrow$X language directions.
For speech translation, we select FLEURS~\citep{conneau2023fleurs}, CoVoST-2~\citep{wang2020covost}, and MuST-C~\citep{di2019must}.
For machine translation, we select FLORES~\citep{costa2022no}, WMT'16~\citep{bojar2016findings}, WMT'19~\citep{barrault2019findings}, and WMT'20~\citep{loic2020findings} corpora.
As a result, we obtain over 592K hypotheses-translation pairs in 11 languages.
The details of dataset statistics are presented in \S\ref{assec:statistics} and Table~\ref{table:statistics_st},~\ref{table:statistics_mt}.

Since the hypotheses-translation data pairs in HypoTranslate dataset are monolingual, we can also use ASR dataset to benefit GenTranslate training, especially for low-resource language pairs.
Relevant studies are illustrated in \S\ref{sssec:roles} and Table~\ref{table:asr_gentrans}. 
Our best result was obtained by first performing translation with SeamlessM4T and then integrating the $N$-best candidates using LLMs.

\begin{table*}[t]
\centering
\resizebox{1.0\textwidth}{!}{
\begin{tabular}{l|ccccccccccccccc|c}
\toprule[1.2pt]
X$\rightarrow$En & Ar & Cy & De & El & Es & Fa & Fr & Hi & It & Ja & Pt & Ta & Uk & Vi & Zh & Avg. \\
\midrule[1.2pt]
\multicolumn{17}{l}{\emph{\textbf{End-to-end ST Methods}}} \\
Whisper-Large V2~\citeyearpar{radford2023robust} & 25.5 & 13.0 & 34.6 & 23.7 & 23.3 & 19.6 & 32.2 & 22.0 & 23.6 & 18.9 & 38.1 & 9.2 & 29.4 & 20.4 & 18.4 & 23.5 \\
AudioPaLM2~\citeyearpar{rubenstein2023audiopalm}* & 29.0 & 7.2 & 38.7 & 18.8 & 26.9 & 25.7 & 36.5 & 21.7 & 27.8 & 11.1 & 38.4 & 15.0 & 26.9 & 15.6 & 21.3 & 24.0 \\
SeamlessM4T-Large~\citeyearpar{barrault2023seamlessm4t} & 32.8 & 31.7 & 35.8 & 25.6 & 25.0 & 28.2 & 33.1 & 26.3 & 25.0 & 17.0 & 38.9 & 16.0 & 30.2 & 21.6 & 19.8 & 27.1 \\
GenTranslate (ours) & \textbf{34.6} & \textbf{33.6} & \textbf{39.2} & \textbf{29.4} & \textbf{29.8} & \textbf{30.5} & \textbf{37.0} & \textbf{28.3} & \textbf{29.7} & \textbf{18.6} & \textbf{43.0} & \textbf{17.4} & \textbf{33.9} & \textbf{24.1} & \textbf{21.7} & \textbf{30.1} \\
\cellhl SeamlessM4T-Large-V2~\citeyearpar{barrault2023seamlessm4tv2}$\bm\dag$ & \cellhl34.7 & \cellhl34.9 & \cellhl37.1 & \cellhl27.3 & \cellhl25.4 & \cellhl30.3 & \cellhl33.7 & \cellhl28.5 & \cellhl26.5 & \cellhl19.5 & \cellhl38.5 & \cellhl22.1 & \cellhl33.2 & \cellhl25.7 & \cellhl23.0 & \cellhl29.4 \\
\cellhl GenTranslate-V2 (ours) & \cellhl\textbf{37.6} & \cellhl\textbf{36.8} & \cellhl\textbf{40.7} & \cellhl\textbf{31.5} & \cellhl\textbf{29.9} & \cellhl\textbf{33.4} & \cellhl\textbf{37.8} & \cellhl\textbf{30.4} & \cellhl\textbf{31.2} & \cellhl\textbf{21.0} & \cellhl\textbf{43.0} & \cellhl\textbf{23.4} & \cellhl\textbf{36.2} & \cellhl\textbf{27.2} & \cellhl\textbf{25.0} & \cellhl\textbf{32.3} \\
\midrule
\multicolumn{17}{l}{\emph{\textbf{Cascaded ASR+MT Methods}}} \\
Whisper + NLLB-3.3b~\citeyearpar{costa2022no} & 35.5 & 29.6 & 40.5 & 31.1 & 30.9 & 28.2 & 39.7 & 26.7 & 30.0 & \underline{24.7} & 44.3 & 20.0 & 35.3 & 26.4 & 25.4 & 31.2 \\
SeamlessM4T (ASR+MT)~\citeyearpar{barrault2023seamlessm4t} & 38.9 & 37.0 & 39.7 & 29.0 & 27.7 & 34.1 & 37.7 & 33.9 & 28.9 & 21.7 & 42.3 & 23.7 & 34.0 & 24.9 & 24.4 & 31.9 \\
GenTranslate (ours) & \textbf{39.9} & \underline{\textbf{39.4}} & \underline{\textbf{41.6}} & \textbf{32.8} & \underline{\textbf{31.2}} & \underline{\textbf{35.9}} & \underline{\textbf{40.6}} & \underline{\textbf{34.9}} & \underline{\textbf{32.1}} & \textbf{22.8} & \underline{\textbf{45.0}} & \textbf{24.1} & \textbf{36.9} & \textbf{27.4} & \textbf{25.7} & \textbf{34.0} \\
\cellhl SeamlessM4T-V2 (ASR+MT)~\citeyearpar{barrault2023seamlessm4tv2}$\bm\dag$ & \cellhl39.2 & \cellhl36.8 & \cellhl39.1 & \cellhl29.4 & \cellhl26.7 & \cellhl33.9 & \cellhl35.7 & \cellhl32.9 & \cellhl29.3 & \cellhl22.5 & \cellhl43.2 & \cellhl25.4 & \cellhl34.8 & \cellhl29.7 & \cellhl25.9 & \cellhl32.3 \\
\cellhl GenTranslate-V2 (ours) & \cellhl\underline{\textbf{40.0}} & \cellhl\textbf{39.1} & \cellhl\textbf{40.9} & \cellhl\underline{\textbf{33.8}} & \cellhl\textbf{30.0} & \cellhl\textbf{35.4} & \cellhl\textbf{40.0} & \cellhl\textbf{33.0} & \cellhl\textbf{31.6} & \cellhl\textbf{23.7} & \cellhl\textbf{44.2} & \cellhl\underline{\textbf{26.4}} & \cellhl\underline{\textbf{37.1}} & \cellhl\underline{\textbf{30.9}} & \cellhl\underline{\textbf{26.9}} & \cellhl\underline{\textbf{34.2}} \\
\bottomrule[1.2pt]
\end{tabular}}
\vspace{-0.1cm}
\caption{Speech translation results on FLEURS \textbf{X$\bm{\rightarrow}$En} test sets in terms of BLEU score, where more results on chrF++ metric~\citep{popovic2017chrf++} are in Table~\ref{table:st_xen_fleurs_chrf}. We use \textbf{bold} to denote surpassing SeamlessM4T baseline, and use \underline{underline} to denote the state-of-the-art. 
The baseline methods are introduced in \S\ref{assec:baselines}. 
* denotes reported by original paper, or else it denotes reproduced by ourselves (same for Table~\ref{table:st_xen_covost2} to~\ref{table:mt_enx_wmt}).
$\bm\dag$ denotes the most latest baseline\protect\footnotemark.
}
\label{table:st_xen_fleurs}
\end{table*}
\footnotetext{Our experiments are mainly conducted on SeamlessM4T-Large as they had already been done before Meta released the latest SeamlessM4T-Large-V2 on November 30th, 2023.
For comprehensive evaluation, we rerun the main experiments on V2, which demonstrate similar effectiveness of our paradigm.
}

\begin{table*}[t]
\centering
\resizebox{1.0\textwidth}{!}{
\begin{tabular}{l|ccccccccccccccc|c}
\toprule[1.2pt]
X$\rightarrow$En & Fr & De & Ca & Es & Ru & Zh & Nl & Tr & Et & Mn & Ar & Lv & Sl & Ja & Id & Avg. \\
\midrule[1.2pt]
\multicolumn{17}{l}{\emph{\textbf{End-to-end ST Methods}}} \\
XLS-R-2b~\citeyearpar{babu2021xls}* & 37.6 & 33.6 & 33.8 & 39.2 & 39.5 & 9.4 & 31.7 & 16.7 & 11.1 & 1.6 & 17.1 & 19.5 & 19.6 & 3.5 & 16.5 & 22.0 \\
Whisper-Large V2~\citeyearpar{radford2023robust} & 35.5 & 35.0 & 31.0 & 39.6 & 42.3 & 16.9 & 40.2 & 27.5 & 14.0 & 0.2 & 38.5 & 13.0 & 16.3 & 24.7 & 47.3 & 28.1 \\
ComSL-Large~\citeyearpar{le2023comsl}* & 38.8 & 36.0 & 35.3 & 40.4 & 49.2 & 21.4 & 39.7 & 33.6 & 19.2 & 2.9 & 41.4 & 21.3 & 31.6 & 21.3 & 46.6 & 31.9 \\
AudioPaLM2~\citeyearpar{rubenstein2023audiopalm}* & \underline{44.8} & \underline{43.4} & 38.4 & \underline{44.2} & \underline{55.6} & \underline{25.5} & \underline{48.3} & \underline{41.0} & \underline{30.0} & 7.6 & 48.7 & \underline{35.0} & \underline{42.6} & 25.9 & 56.2 & \underline{39.1} \\
SeamlessM4T-Large~\citeyearpar{barrault2023seamlessm4t} & 41.3 & 38.8 & 38.4 & 41.1 & 48.6 & 20.9 & 41.1 & 31.2 & 26.3 & 7.5 & 45.0 & 26.5 & 37.6 & 21.8 & 51.4 & 34.5 \\
GenTranslate (ours) & \textbf{41.7} & \textbf{39.2} & \textbf{38.7} & \textbf{42.0} & \textbf{50.1} & \textbf{21.6} & \textbf{42.1} & \textbf{33.5} & \textbf{28.2} & \textbf{8.7} & \underline{\textbf{49.7}} & \textbf{30.3} & \textbf{38.2} & \textbf{22.9} & \textbf{54.3} & \textbf{36.1} \\
\cellhl SeamlessM4T-Large-V2~\citeyearpar{barrault2023seamlessm4tv2} & \cellhl42.4 & \cellhl40.0 & \cellhl39.0 & \cellhl42.9 & \cellhl53.6 & \cellhl22.4 & \cellhl42.7 & \cellhl33.2 & \cellhl26.9 & \cellhl8.6 & \cellhl46.5 & \cellhl27.5 & \cellhl41.7 & \cellhl23.7 & \cellhl52.6 & \cellhl36.2 \\
\cellhl GenTranslate-V2 (ours) & \cellhl\textbf{42.7} & \cellhl\textbf{40.6} & \cellhl\textbf{39.4} & \cellhl\textbf{43.6} & \cellhl\textbf{54.0} & \cellhl\textbf{23.3} & \cellhl\textbf{44.8} & \cellhl\textbf{37.0} & \cellhl\textbf{27.7} & \cellhl\textbf{10.2} & \cellhl\textbf{48.0} & \cellhl\textbf{30.5} & \cellhl\textbf{42.3} & \cellhl\textbf{25.4} & \cellhl\textbf{55.9} & \cellhl\textbf{37.7} \\
\midrule
\multicolumn{17}{l}{\emph{\textbf{Cascaded ASR+MT Methods}}} \\
Whisper + NLLB-3.3b~\citeyearpar{costa2022no} & 34.4 & 35.5 & 31.7 & 37.9 & 45.4 & 19.0 & 39.8 & 26.7 & 17.5 & 0.1 & 37.0 & 20.6 & 29.4 & 25.5 & 45.9 & 29.8 \\
Whisper + mBART-50~\citeyearpar{le2023comsl}* & 38.8 & 37.0 & 33.0 & 40.7 & 49.0 & 21.5 & 39.9 & 32.7 & 16.3 & 0.4 & 37.0 & 21.4 & 25.0 & 23.0 & 45.5 & 30.7 \\
SeamlessM4T (ASR+MT)~\citeyearpar{barrault2023seamlessm4t} & 41.5 & 39.8 & 37.5 & 41.1 & 53.2 & 21.4 & 42.4 & 29.9 & 26.5 & 8.0 & 45.2 & 28.8 & 38.6 & 22.0 & 50.6 & 35.1 \\
GenTranslate (ours) & \textbf{41.8} & \textbf{40.2} & \textbf{38.4} & \textbf{42.1} & \textbf{53.7} & \textbf{22.9} & \textbf{43.8} & \textbf{34.3} & \textbf{29.4} & \textbf{9.5} & \underline{\textbf{49.7}} & \textbf{31.2} & \textbf{39.6} & \textbf{22.3} & \textbf{54.6} & \textbf{36.9} \\
\cellhl SeamlessM4T-V2 (ASR+MT)~\citeyearpar{barrault2023seamlessm4tv2} & \cellhl43.0 & \cellhl40.6 & \cellhl38.8 & \cellhl43.0 & \cellhl55.2 & \cellhl22.9 & \cellhl43.2 & \cellhl33.9 & \cellhl27.2 & \cellhl8.6 & \cellhl47.0 & \cellhl27.8 & \cellhl41.9 & \cellhl24.7 & \cellhl53.1 & \cellhl36.7 \\
\cellhl GenTranslate-V2 (ours) & \cellhl\textbf{43.1} & \cellhl\textbf{41.1} & \cellhl\underline{\textbf{39.5}} & \cellhl\textbf{43.3} & \cellhl\underline{\textbf{55.6}} & \cellhl\textbf{24.5} & \cellhl\textbf{44.9} & \cellhl\textbf{37.4} & \cellhl\textbf{27.8} & \cellhl\underline{\textbf{10.3}} & \cellhl\textbf{48.7} & \cellhl\textbf{30.4} & \cellhl\textbf{42.0} & \cellhl\underline{\textbf{26.0}} & \cellhl\underline{\textbf{58.4}} & \cellhl\textbf{38.2} \\
\bottomrule[1.2pt]
\end{tabular}}
\vspace{-0.1cm}
\caption{Speech translation results on CoVoST-2 \textbf{X$\bm{\rightarrow}$En} test sets in terms of BLEU score.
Remarks follow Table~\ref{table:st_xen_fleurs}.
}
\label{table:st_xen_covost2}
\vspace{-0.4cm}
\end{table*}

\section{Experiments}
\label{sec:exp}

\subsection{Setup}
\label{ssec:setup}

\subsubsection{Model Selection}
\label{sssec:models}

\noindent\textbf{LLMs.}
We select the popular LLaMA-2~\citep{touvron2023llama2} for our paradigm.
Specifically, we employ LLaMA-2-7b\footnote{\url{https://huggingface.co/meta-llama/Llama-2-7b-hf}} for English-target directions (X$\rightarrow$En) and LLaMA-2-13b for non-English-target directions (En$\rightarrow$X), as LLaMA-2 shows superior ability on English language while less-optimal on other languages.
In addition, for En$\rightarrow$X we also try some latest multilingual LLMs like BigTranslate\footnote{\url{https://huggingface.co/James-WYang/BigTranslate}\label{fn5}}~\citep{yang2023bigtrans} and ALMA\footnote{\url{https://huggingface.co/haoranxu/ALMA-13B}\label{fn6}}~\citep{xu2023paradigm} that are finetuned on LLaMA-13b.

\noindent\textbf{Adapter.}
We follow the default settings of LLaMA-Adapter~\citep{zhang2023llama}.
The number of tunable Transformer layers $L$ is set to $H-1$, which means all layers except the first one are tunable with inserted prompts.
The prompt length $U$ is set to 10.
More details are provided in \S\ref{assec:model_setups}.

\subsubsection{Training Details}
\label{sssec:training}
The batch size is set to 4, with accumulation iterations set to 8 (\emph{i.e.}, real batch size is 32).
We train 2 epochs with AdamW optimizer~\citep{loshchilov2018decoupled}, with learning rate initialized to $1e^{-2}$ and then linearly decrease to $1e^{-5}$ during training.

\begin{table*}[t]
\centering
\resizebox{1.0\textwidth}{!}{
\begin{tabular}{l|ccccccc|cccc|cccc}
\toprule[1.2pt]
\multirow{2}{*}{En$\rightarrow$X} & \multicolumn{7}{c|}{FLEURS} & \multicolumn{4}{c|}{CoVoST-2} & \multicolumn{4}{c}{MuST-C} \\
& Es & Fr & It & Ja & Pt & Zh & Avg. & Fa & Ja & Zh & Avg. & Es & It & Zh & Avg. \\
\midrule[1.2pt]
\multicolumn{16}{l}{\emph{\textbf{End-to-end ST Methods}}} \\
SeamlessM4T-Large~\citeyearpar{barrault2023seamlessm4t} & 23.8 & 41.6 & 23.9 & 21.0 & 40.8 & 28.6 & 30.0 & 18.3 & 24.0 & 34.1 & 25.5 & \textbf{34.2} & \textbf{29.9} & 16.2 & 26.8 \\
GenTranslate (ours) & \textbf{25.4} & \textbf{43.1} & \textbf{25.5} & \textbf{28.3} & \textbf{42.4} & \textbf{34.3} & \textbf{33.2} & \textbf{21.1} & \textbf{29.1} & \textbf{42.8} & \textbf{31.0} & 33.9 & 29.4 & \textbf{18.5} & \textbf{27.3} \\
\cellhl SeamlessM4T-Large-V2~\citeyearpar{barrault2023seamlessm4tv2} & \cellhl23.8 & \cellhl42.6 & \cellhl24.5 & \cellhl21.7 & \cellhl43.0 & \cellhl29.5 & \cellhl30.9 & \cellhl16.9 & \cellhl23.5 & \cellhl34.6 & \cellhl25.0 & \cellhl32.1 & \cellhl\textbf{27.5} & \cellhl15.6 & \cellhl25.1 \\
\cellhl GenTranslate-V2 (ours) & \cellhl\textbf{25.5} & \cellhl\textbf{44.0} & \cellhl\textbf{26.3} & \cellhl\textbf{28.9} & \cellhl\underline{\textbf{44.5}} & \cellhl\textbf{34.9} & \cellhl\textbf{34.0} & \cellhl\textbf{19.4} & \cellhl\textbf{29.0} & \cellhl\underline{\textbf{43.6}} & \cellhl\textbf{30.7} & \cellhl\textbf{32.2} & \cellhl27.3 & \cellhl\textbf{18.1} & \cellhl\textbf{25.9} \\
\midrule
\multicolumn{16}{l}{\emph{\textbf{Cascaded ASR+MT Methods}}} \\
Whisper + NLLB-3.3b~\citeyearpar{costa2022no} & 25.1 & 41.3 & 25.0 & 19.0 & 41.5 & 23.5 & 29.2 & 13.6 & 19.0 & 32.0 & 21.5 & 35.3 & 29.9 & 13.5 & 26.2 \\
SeamlessM4T-Large (ASR+MT)~\citeyearpar{barrault2023seamlessm4t} & 24.6 & 44.6 & 25.4 & 22.5 & 41.9 & 31.2 & 31.7 & 18.8 & 24.0 & 35.1 & 26.0 & 35.1 & 30.8 & 17.7 & 27.9 \\
GenTranslate (ours) & \textbf{26.8} & \underline{\textbf{45.0}} & \underline{\textbf{26.6}} & \underline{\textbf{29.4}} & \textbf{43.1} & \underline{\textbf{36.8}} & \underline{\textbf{34.6}} & \underline{\textbf{21.8}} & \underline{\textbf{30.5}} & \textbf{43.3} & \underline{\textbf{31.9}} & \underline{\textbf{35.5}} & \underline{\textbf{31.0}} & \underline{\textbf{19.6}} & \underline{\textbf{28.7}} \\
\cellhl SeamlessM4T-V2 (ASR+MT)~\citeyearpar{barrault2023seamlessm4tv2} & \cellhl24.7 & \cellhl44.1 & \cellhl25.1 & \cellhl20.6 & \cellhl43.6 & \cellhl30.6 & \cellhl31.5 & \cellhl17.4 & \cellhl23.8 & \cellhl35.4 & \cellhl25.5 & \cellhl33.0 & \cellhl27.8 & \cellhl14.5 & \cellhl25.1 \\
\cellhl GenTranslate-V2 (ours) & \cellhl\underline{\textbf{27.0}} & \cellhl\textbf{44.3} & \cellhl\textbf{26.4} & \cellhl\textbf{27.8} & \cellhl\underline{\textbf{44.5}} & \cellhl\textbf{36.1} & \cellhl\textbf{34.4} & \cellhl\textbf{20.8} & \cellhl\textbf{29.7} & \cellhl\textbf{43.5} & \cellhl\textbf{31.3} & \cellhl\textbf{33.2} & \cellhl\textbf{28.3} & \cellhl\textbf{16.9} & \cellhl\textbf{26.1} \\
\bottomrule[1.2pt]
\end{tabular}}
\vspace{-0.1cm}
\caption{Speech translation results on FLEURS, CoVoST-2, and MuST-C \textbf{En$\bm{\rightarrow}$X} test sets in terms of BLEU score.
We use \textbf{bold} to highlight surpassing SeamlessM4T baseline, and use \underline{underline} to highlight the state-of-the-art performance.
The baseline methods are introduced in \S\ref{assec:baselines}, and all of their results are reproduced by ourselves.
}
\label{table:st_enx}
\vspace{-0.2cm}
\end{table*}

\begin{table*}[t]
\centering
\resizebox{1.0\textwidth}{!}{
\begin{tabular}{l|cccccccccc|c}
\toprule[1.2pt]
X$\rightarrow$En & Ar & De & El & Es & Fa & Fr & It & Ja & Uk & Zh & Avg. \\
\midrule[1.2pt]
ALMA-13b~\citep{xu2023paradigm} & 10.8 & 27.7 & 12.1 & 18.1 & 10.2 & 27.4 & 19.6 & 14.2 & 22.7 & 16.9 & 18.0 \\
BigTranslate~\citep{yang2023bigtrans} & 18.6 & 35.9 & 9.5 & 29.0 & 1.4 & 38.7 & 29.0 & 16.9 & 25.9 & 23.0 & 22.8 \\
NLLB-3.3b~\citep{costa2022no} & 43.0 & 44.6 & 37.7 & 32.2 & 38.7 & 46.2 & 34.6 & \underline{28.1} & 40.8 & 29.5 & 37.5 \\
SeamlessM4T-Large~\citep{barrault2023seamlessm4t} & 43.7 & 45.1 & 37.7 & 31.5 & 39.0 & 45.1 & 35.2 & 26.1 & 41.2 & 29.9 & 37.5 \\
GenTranslate (ours) & \underline{\textbf{43.9}} & \underline{\textbf{45.3}} & \underline{\textbf{38.5}} & \underline{\textbf{35.5}} & \underline{\textbf{39.4}} & \textbf{46.4} & \underline{\textbf{36.6}} & \textbf{26.7} & \underline{\textbf{41.8}} & \underline{\textbf{30.5}} & \underline{\textbf{38.5}} \\
\cellhl SeamlessM4T-Large-V2~\citep{barrault2023seamlessm4tv2} & \cellhl41.5 & \cellhl44.1 & \cellhl35.6 & \cellhl29.9 & \cellhl37.6 & \cellhl45.5 & \cellhl33.5 & \cellhl25.5 & \cellhl39.0 & \cellhl29.0 & \cellhl36.1 \\
\cellhl GenTranslate-V2 (ours) & \cellhl\textbf{42.0} & \cellhl\textbf{44.5} & \cellhl\textbf{36.6} & \cellhl\textbf{34.4} & \cellhl\textbf{38.1} & \cellhl\underline{\textbf{46.7}} & \cellhl\textbf{35.1} & \cellhl\textbf{26.7} & \cellhl\textbf{39.3} & \cellhl\textbf{29.9} & \cellhl\textbf{37.3} \\
\bottomrule[1.2pt]
\end{tabular}}
\vspace{-0.1cm}
\caption{Machine translation results on FLORES \textbf{X$\bm{\rightarrow}$En} test sets in terms of BLEU score.
Remarks follow Table~\ref{table:st_enx}.
}
\label{table:mt_xen_flores}
\vspace{-0.4cm}
\end{table*}

\begin{table}[t]
\centering
\resizebox{1.0\columnwidth}{!}{
\begin{tabular}{p{4cm}|c|cc|cc|c}
\toprule[1.2pt]
\multirow{2}{*}{En$\rightarrow$X} & WMT'16 & \multicolumn{2}{c|}{WMT'19} & \multicolumn{2}{c|}{WMT'20} & \multirow{2}{*}{Avg.} \\
& Ro & Cs & Lt & Ja & Zh & \\
\midrule[1.2pt]
ALMA-13b~\citeyearpar{xu2023paradigm} & 6.2 & 6.1 & 0.3 & 3.5 & 11.3 & 5.5 \\
BigTranslate~\citeyearpar{yang2023bigtrans} & 21.4 & 19.0 & 8.7 & 7.3 & 29.0 & 17.1 \\
NLLB-3.3b~\citeyearpar{costa2022no} & 31.0 & 25.3 & 16.0 & 15.2 & 26.9 & 22.9 \\
SeamlessM4T-Large & 32.7 & 26.0 & 17.2 & 17.0 & 27.2 & 24.0 \\
GenTranslate (ours) & \underline{\textbf{33.5}} & \underline{\textbf{27.2}} & \underline{\textbf{19.4}} & \underline{\textbf{21.4}} & \textbf{30.7} & \underline{\textbf{26.4}} \\
\cellhl SeamlessM4T-Large-V2 & \cellhl32.2 & \cellhl25.2 & \cellhl16.2 & \cellhl15.2 & \cellhl28.7 & \cellhl23.5 \\
\cellhl GenTranslate-V2 (ours) & \cellhl\textbf{33.2} & \cellhl\textbf{26.6} & \cellhl\textbf{18.2} & \cellhl\textbf{19.3} & \cellhl\underline{\textbf{31.6}} & \cellhl\textbf{25.8} \\
\bottomrule[1.2pt]
\end{tabular}}
\vspace{-0.1cm}
\caption{Machine translation results on WMT'{16,19,20} \textbf{En$\bm{\rightarrow}$X} test sets in BLEU.
Remarks follow Table~\ref{table:st_enx}.
}
\label{table:mt_enx_wmt}
\vspace{-0.2cm}
\end{table}

\subsection{Comparison with the State-of-the-art}
\label{ssec:compare_with_sota}

\subsubsection{Speech Translation}
\label{sssec:st}

\noindent\textbf{X$\rightarrow$English (En).}
Table~\ref{table:st_xen_fleurs} and \ref{table:st_xen_covost2} present the X$\rightarrow$En speech translation performance on FLEURS and CoVoST-2 datasets.
We can observe from Table~\ref{table:st_xen_fleurs} that all the strong baselines like Whisper, AudioPaLM2 and SeamlessM4T-Large perform well on 15 X$\rightarrow$En directions, where SeamlessM4T-Large is the best (27.1 BLEU).
With LLMs introduced for $N$-best integration, our GenTranslate achieves consistent improvements on various source languages X, where further analysis on language family is presented in \S\ref{sssec:analysis_lang_family}.
As a result, our GenTranslate shows 3.0 BLEU improvement over SeamlessM4T-Large, which verifies the effectiveness of LLMs for generative translation\footnote{Latest SeamlessM4T-Large-V2 achieves significant gains over V1, based on which the proposed GenTranslate also shows similar effectiveness in our study.}.

Following the speech translation literature, we also investigate cascaded ASR+MT methods for evaluation.
We can observe from Table~\ref{table:st_xen_fleurs} that, with the same SeamlessM4T-Large backbone, cascaded system outperforms end-to-end system by 4.8 BLEU score, which is consistent with previous findings~\citep{xu2023recent}.
Latest SeamlessM4T-Large-V2 further improves V1 model, and our GenTranslate shows significant and consistent gains of performance over theses two backbones.

Table~\ref{table:st_xen_covost2} presents the X$\rightarrow$En ST results on more language directions of CoVoST-2 dataset, where we introduce more latest baselines for comprehensive comparison.
In end-to-end methods, SeamlessM4T-Large achieves a good 34.5 BLEU score though underperforms the state-of-the-art AudioPaLM2\footnote{We speculate it could be attributed to the train-test domain mismatch because SeamlessM4T-Large outperforms AudioPaLM2 by a large margin on FLEURS dataset in Table~\ref{table:st_xen_fleurs}.}.
In comparison, our GenTranslate achieves a promising improvement over SeamlessM4T.
Similar phenomenon can be observed in cascaded systems, where SeamlessM4T significantly outperforms the competitive baselines that combine state-of-the-art ASR and MT models, and our GenTranslate moves one step forward with 1.8 BLEU improvement.
Similar improvements can be observed on SeamlessM4T-Large-V2 backbone.

\begin{table*}[t]
\centering
\resizebox{1.0\textwidth}{!}{
\begin{tabular}{l|ccccc|cccc|cccccc}
\toprule[1.2pt]
\multirow{2}{*}{En$\rightarrow$X} & \multicolumn{5}{c|}{FLEURS} & \multicolumn{4}{c|}{CoVoST-2} & \multicolumn{6}{c}{WMT} \\
& Es & Fr & It & Pt & Avg. & Fa & Ja & Zh & Avg. & Ro & Cs & It & Ja & Zh & Avg. \\
\midrule[1.2pt]
SeamlessM4T-Large~\citeyearpar{barrault2023seamlessm4t} & 24.6 & 44.6 & 25.4 & 41.9 & 34.1 & 18.8 & 24.0 & 35.1 & 26.0 & 32.7 & 26.0 & 17.2 & 17.0 & 27.2 & 24.0 \\
\midrule
\multicolumn{16}{l}{GenTranslate \emph{with}} \\
\quad BigTranslate~\citeyearpar{yang2023bigtrans} & 25.3 & 44.2 & 25.5 & 40.8 & 34.0 & 5.2 & 23.5 & 42.6 & 23.8 & 31.3 & 24.9 & 15.8 & 13.9 & 27.9 & 22.8 \\
\quad ALMA-13b~\citeyearpar{xu2023paradigm} & 24.9 & 43.5 & 25.1 & 40.6 & 33.5 & 19.2 & 29.3 & \textbf{43.9} & 30.8 & 31.1 & 25.5 & 17.7 & 17.3 & 26.8 & 23.7 \\
\quad LLaMA-2-13b~\citeyearpar{touvron2023llama2} & \textbf{26.8} & \textbf{45.0} & \textbf{26.6} & \textbf{43.1} & \textbf{35.4} & \textbf{21.8} & \textbf{30.5} & 43.3 & \textbf{31.9} & \textbf{33.5} & \textbf{27.2} & \textbf{19.4} & \textbf{21.4} & \textbf{30.7} & \textbf{26.4} \\
\bottomrule[1.2pt]
\end{tabular}}
\vspace{-0.1cm}
\caption{Effect of different multilingual LLMs on GenTranslate, in terms of the speech translation results on FLEURS and CoVoST-2 En$\bm{\rightarrow}$X test sets, as well as the machine translation results on WMT En$\bm{\rightarrow}$X test sets.
}
\label{table:ablation_llm}
\vspace{-0.4cm}
\end{table*}

\begin{table}[t]
\centering
\resizebox{1.0\columnwidth}{!}{
\begin{tabular}{l|c}
\toprule[1.2pt]
De$\rightarrow$En & BLEU Score \\
\midrule[1.2pt]
\multicolumn{2}{l}{\emph{\textbf{End-to-end ST Methods}}} \\
SeamlessM4T (ST)~\citep{barrault2023seamlessm4t} & 35.8 \\
SeamlessM4T (ST) + GenTranslate & \textbf{39.2} \\
\midrule
\multicolumn{2}{l}{\emph{\textbf{Cascaded ASR+MT Methods}}} \\
SeamlessM4T (ASR+MT)~\citep{barrault2023seamlessm4t} & 39.7 \\
SeamlessM4T (ASR+MT) + GenTranslate & \underline{\textbf{41.6}} \\
\midrule
\multicolumn{2}{l}{\emph{\textbf{ASR+GenTranslate Method}}} \\
SeamlessM4T (ASR) + GenTranslate \emph{with} & \\
\quad LLaMA-2-7b~\citep{touvron2023llama2} & 36.8 \\
\quad BigTranslate~\citep{yang2023bigtrans} & 38.2 \\
\quad ALMA-7b~\citep{xu2023paradigm} & \textbf{40.6} \\
\bottomrule[1.2pt]
\end{tabular}}
\vspace{-0.1cm}
\caption{Performance of ASR+GenTranslate system on FLEURS De$\rightarrow$En ST test set.
As shown in Fig.~\ref{fig4}, it first uses ASR to produce German $N$-best hypotheses, and then leverages LLMs to generate the English translation from them.
Different LLMs are investigated here.
}
\label{table:asr_gentrans}
\vspace{-0.4cm}
\end{table}

\vspace{0.1cm}
\noindent\textbf{English (En)$\rightarrow$X.}
For comprehensive evaluation, we also present En$\rightarrow$X ST results on three datasets in Table~\ref{table:st_enx}.
SeamlessM4T (both Large and Large-V2) achieves excellent performance on En$\rightarrow$X ST tasks under both end-to-end and cascaded systems.
In comparison, our proposed GenTranslate achieves significant performance improvements ($\sim$3 BLEU score) in various language directions.
Since En$\rightarrow$X translation tasks produce non-English $N$-best hypotheses for LLM integration, such performance gains indicates the excellent multilingual abilities of LLMs (\emph{i.e.}, LLaMA-2).

\subsubsection{Machine Translation}
\label{sssec:mt}

\noindent\textbf{X$\rightarrow$English (En).}
Table~\ref{table:mt_xen_flores} presents the X$\rightarrow$En MT results on FLORES dataset.
The baseline methods ALMA-13b and BigTranslate show limited performance.
NLLB-3.3b achieves an improved performance of 37.5 BLEU, which is comparable to SeamlessM4T-Large.
Based on that, our GenTranslate achieves the state-of-the-art with consistent gains on all language directions except Ja$\rightarrow$En.

\vspace{0.1cm}
\noindent\textbf{English (En)$\rightarrow$X.}
Table~\ref{table:mt_enx_wmt} presents the En$\rightarrow$X MT results on WMT test sets.
Similar to previous results, we observe much higher BLEU scores of NLLB-3.3b than ALMA-13b and BigTranslate.
SeamlessM4T-Large surpasses NLLB-3.3b by large-scale multitask training.
The proposed GenTranslate achieves the state-of-the-arts on all language directions with a gain of 2.4 BLEU score.
Please note that SeamlessM4T-Large-V2 underperforms V1 on selected MT datasets, but our GenTranslate achieves consistent gains on both of them.

\begin{figure}[t]
\begin{center}
\includegraphics[width=1.0\columnwidth]{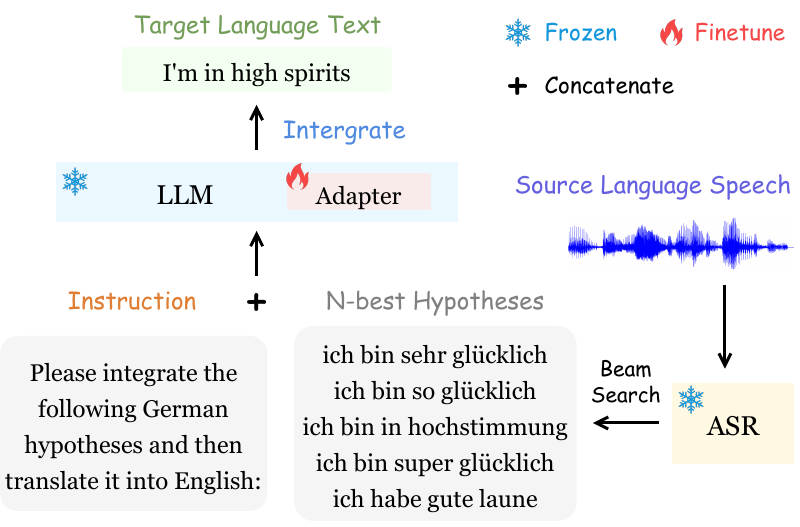}
\end{center}
\vspace{-0.2cm}
\caption{Illustration of the ``ASR+GenTranslate'' system for ST task as introduced in Table~\ref{table:asr_gentrans} and \S\ref{sssec:roles}.
This system engages LLMs into the translation process by combining it with the $N$-best integration process.
}
\vspace{-0.4cm}
\label{fig4}
\end{figure}

In summary, we observe consistent improvements of GenTranslate over various baselines (\emph{i.e.}, SeamlessM4T, Whisper, etc.), various tasks (\emph{i.e.}, ST and MT), various test data (\emph{i.e.}, FLEURS, WMT, etc.), and various language directions (\emph{i.e.}, X$\rightarrow$En and En$\rightarrow$X).
Therefore, the effectiveness and generality of our approach are well verified.

\subsection{Ablation Study}
\label{ssec:ablation}

\subsubsection{Effect of Different LLMs}
\label{sssec:ablation_llm}
According to Table~\ref{table:st_enx} and \ref{table:mt_enx_wmt}, LLaMA-2 has shown excellent multilingual ability.
To further investigate the role of this ability in GenTranslate, we select two latest multilingual LLMs for comparison, \emph{i.e.}, BigTranslate and ALMA-13b.
Table~\ref{table:ablation_llm} shows that both of them perform worse than LLaMA-2-13b for ST and MT tasks.
One explanation is, BigTranslate and ALMA-13b are finetuned on MT task that requires cross-lingual ability, while the En$\rightarrow$X GenTranslate mainly requires strong monolingual ability of language X, such mismatch may explain why MT finetuning fails to enhance GenTranslate. 

\begin{table*}[t]
\centering
\resizebox{1.0\textwidth}{!}{
\begin{tabular}{l|ccccccccccccccc|c}
\toprule[1.2pt]
X$\rightarrow$En & Ar & Cy & De & El & Es & Fa & Fr & Hi & It & Ja & Pt & Ta & Uk & Vi & Zh & Avg. \\
\midrule[1.2pt]
SeamlessM4T (ASR+MT) & 38.9 & 37.0 & 39.7 & 29.0 & 27.7 & 34.1 & 37.7 & 33.9 & 28.9 & 21.7 & 42.3 & 23.7 & 34.0 & 24.9 & 24.4 & 31.9 \\
\midrule
\multicolumn{17}{l}{GenTranslate \emph{with}} \\
\quad LLaMA-Adapter & {{39.9}} & {\textbf{39.4}} & {{41.6}} & {\textbf{32.8}} & {{31.2}} & {{35.9}} & {\textbf{40.6}} & {{34.9}} & {{32.1}} & \textbf{22.8} & {{45.0}} & {\textbf{24.1}} & {\textbf{36.9}} & {\textbf{27.4}} & {{25.7}} & {{34.0}} \\
\quad LLaMA-LoRA & \textbf{40.2} & 39.3 & \textbf{41.8} & \textbf{32.8} & \textbf{31.6} & \textbf{36.0} & \textbf{40.6} & \textbf{35.2} & \textbf{32.4} & 22.5 & \textbf{45.1} & \textbf{24.1} & 36.7 & 27.1 & \textbf{26.0} & \textbf{34.1} \\
\bottomrule[1.2pt]
\end{tabular}}
\vspace{-0.1cm}
\caption{Comparison between LLaMA-Adapter and LLaMA-LoRA for efficient LLM finetuning in our GenTranslate, in terms of the speech translation results on FLEURS X$\rightarrow$En test sets.
}
\label{table:adapter_vs_lora}
\end{table*}

\begin{table*}[t]
\centering
\resizebox{1.0\textwidth}{!}{
\begin{tabular}{l|ccccccccccc|cccccc}
\toprule[1.2pt]
\multirow{2}{*}{X$\rightarrow$En} & \multicolumn{11}{c|}{Indo-European} & \multicolumn{6}{c}{non-Indo-European} \\
& Fa & Hi & It & Es & Fr & Pt & Cy & De & El & Uk & Avg. & Ar & Vi & Ja & Ta & Zh & Avg. \\
\midrule[1.2pt]
SeamlessM4T (ASR+MT) & 34.1 & 33.9 & 28.9 & 27.7 & 37.7 & 42.3 & 37.0 & 39.7 & 29.0 & 34.0 & 34.4 & 38.9 & 24.9 & 21.7 & 23.7 & 24.4 & 26.7 \\
GenTranslate (ours) & 35.9 & 34.9 & 32.1 & 31.2 & 40.6 & 45.0 & 39.4 & 41.6 & 32.8 & 36.9 & 37.0 & 39.9 & 27.4 & 22.8 & 24.1 & 25.7 & 28.0 \\
\midrule
$\Delta$ BLEU & 1.8 & 1.0 & 3.2 & 3.5 & 2.9 & 2.7 & 2.4 & 1.9 & 3.8 & 2.9 & 2.6 & 1.0 & 2.5 & 1.1 & 0.4 & 1.4 & 1.3 \\
\bottomrule[1.2pt]
\end{tabular}}
\vspace{-0.1cm}
\caption{Effect of language family on our proposed GenTranslate. 
We report speech translation results on FLEURS X$\rightarrow$En test sets in this study.
For simplicity, we split all the languages into two families, \emph{i.e.}, Indo-European (same as English) and non-Indo-European, and more detailed information are presented in Table~\ref{table:detailed_family_info}.
}
\vspace{-0.2cm}
\label{table:analysis_language_family}
\end{table*}

\begin{table}[t]
\centering
\resizebox{1.0\columnwidth}{!}{
\begin{tabular}{cc|cccccc|c}
\toprule[1.2pt]
\multicolumn{2}{l|}{X$\rightarrow$En} & Ar & De & Es & Fr & Pt & Zh & Avg. \\
\midrule[1.2pt]
\multicolumn{2}{l|}{SeamlessM4T-Large} & 32.8 & 35.8 & 25.0 & 33.1 & 38.9 & 19.8 & 30.9 \\
\midrule
\multicolumn{9}{l}{GenTranslate \emph{with}} \\
\multirow{6}{*}{N =} & 1 & 31.3 & 35.4 & 26.9 & 35.2 & 41.5 & 19.3 & 31.6 \\
& 3 & 34.2 & 38.9 & 29.5 & 36.4 & 42.8 & 21.3 & 33.9 \\
& 5 & 34.6 & 39.2 & \textbf{29.8} & \textbf{37.0} & 43.0 & \textbf{21.7} & 34.2 \\
& 8 & 34.8 & \textbf{39.9} & 29.4 & 36.9 & 43.0 & 21.5 & \textbf{34.3} \\
& 10 & \textbf{35.3} & 39.8 & 29.4 & 36.6 & \textbf{43.2} & 21.6 & \textbf{34.3} \\
& 15 & 34.9 & 39.5 & 29.6 & 36.4 & 42.8 & 21.6 & 34.1 \\
\bottomrule[1.2pt]
\end{tabular}}
\vspace{-0.1cm}
\caption{Effect of $N$-best list size on GenTranslate (default N=5), in terms of ST results on FLEURS X$\bm{\rightarrow}$En.
}
\label{table:ablation_nbest}
\vspace{-0.3cm}
\end{table}

\subsubsection{Role of LLMs in GenTranslate}
\label{sssec:roles}
To further investigate the role of LLMs in our GenTranslate, we build an ASR+GenTranslate system for ST task as shown in Fig.~\ref{fig4}.
Take De$\rightarrow$En as an example, we first send the German speech input into ASR to produce $N$-best transcriptions, which are then fed by LLMs to generate English translation.
In other words, LLMs are assigned $N$-best integration and translation tasks at the same time.
As shown in Table~\ref{table:asr_gentrans}, among the three evaluated LLMs, ALMA-7b achieves the best performance thanks to its MT finetuning during development, but it still underperforms the best cascaded method (40.6 vs. 41.6).
We can conclude from such observations that 1) LLaMA-2 provides reasonable  translation ability and it can be further improved via MT task finetuning (\emph{i.e.}, ALMA). 2) In this study, LLM underperforms SeamlessM4T in translation task, but it shows remarkable ability in $N$-best integration. 
Therefore, future work may focus on how to better engage LLMs into the translation part.

\subsubsection{Effect of $N$-best List Size}
\label{sssec:ablation_nbest}
GenTranslate relies on powerful LLMs and informative $N$-best hypotheses to generate higher-quality translation output.
Therefore, the amount of information in $N$-best hypotheses could be a key factor of GenTranslate's performance.
We can observe from Table~\ref{table:ablation_nbest} that with the increase of N, the performance of GenTranslate first improves and then drops, where the best choice ranges from 5 to 10.
We believe that small N results in insufficient information for generation of ground-truth translation, while too large N leads to information redundancy and thus increases the miscorrection and hallucination.
In this work, we set N to 5 for the best trade-off between efficiency and quality.

\subsubsection{LLaMA-Adapter vs. LLaMA-LoRA}
\label{sssec:adapter_vs_lora}
Apart from LLaMA-Adapter, low-rank adaptation (LoRA)~\citep{hu2021lora, yu2023low} is another popular efficient LLM finetuning strategy.
Table~\ref{table:adapter_vs_lora} compares the performance between LLaMA-Adapter and LLaMA-LoRA for proposed GenTranslate, in terms of the BLEU results of ST task on FLEURS X$\rightarrow$En test sets.
We can observe similar BLEU performance of these two strategies on GenTranslate (34.0 vs. 34.1), indicating that the efficient LLM finetuning strategy is not a key factor in GenTranslate paradigm.

\begin{table*}[t]
\centering
\resizebox{0.93\textwidth}{!}{
\begin{tabular}{c|l|c}
\toprule[1.2pt]
Method & Utterance & BLEU Score \\
\midrule[1.2pt]
\multirow{5}{*}{$N$-best Candidates} & \textcolor{red}{TV} reports show that white smoke is \textcolor{red}{escaping} from the plant. & 28.6 \\
& \textcolor{red}{TV} reports show that white smoke is \textcolor{red}{escaping} from the facility. & 12.2 \\
& \textcolor{teal}{Television} reports show that white smoke is \textcolor{red}{escaping} from the plant. & 34.2 \\
& \textcolor{teal}{Television} reports show that white smoke is \textcolor{red}{escaping} from the facility. & 19.2 \\
& \textcolor{red}{TV} reports show that white smoke \textcolor{red}{escapes} from the plant. & 31.7 \\
\midrule
GenTranslate (ours) & \textcolor{teal}{Television} reports show white smoke \textcolor{teal}{coming} out of the plant. & \textbf{58.8} \\
\midrule
Ground-truth Translation & Television reports show white smoke coming from the plant. & - \\
\bottomrule[1.2pt]
\end{tabular}}
\vspace{-0.1cm}
\caption{Case study of GenTranslate.
The test sample is selected from the FLEURS De$\rightarrow$En ST test set.
}
\label{table:case_study}
\vspace{-0.2cm}
\end{table*}

\subsection{Analysis}
\label{ssec:analysis}

\subsubsection{Effect of Language Family}
\label{sssec:analysis_lang_family}
Table~\ref{table:analysis_language_family} analyzes the effect of language family using the X$\rightarrow$En ST results.
The source language X is grouped into two categories depending on whether it belongs to Indo-European family (English is also Indo-European language).
First, we observe better results of SeamlessM4T when X belongs to Indo-European family, indicating that translation within same family is easier than across different families.
Then, we also observe larger BLEU improvement of GenTranslate over baseline when X is Indo-European language (2.6 vs. 1.3).
The reason could be, within-family translation produces $N$-best hypotheses with higher quality and richer information, which is beneficial to GenTranslate.

\subsubsection{Case Study}
\label{sssec:case_study}
Table~\ref{table:case_study} shows a case study where GenTranslate outperforms the 1-best hypothesis by a large margin.
We may speculate two key points about its working mechanism, where it first extract the word ``Television'' from $3^{rd}/4^{th}$ hypotheses to replace ``TV'' and then reason out the word ``coming'' that does not exist in $N$-best list.
Therefore, our paradigm may not only integrate the $N$-best sentences for better result, but also improve the translation quality by itself.
Another non-English case study is in Appendix~\ref{assec:supply_case_study}.

\subsubsection{Visualizations of GenTranslate Output}
\label{sssec:visualizations}
\begin{figure}[t]
\begin{center}
\vspace{0.1cm}
\includegraphics[width=0.97\columnwidth]{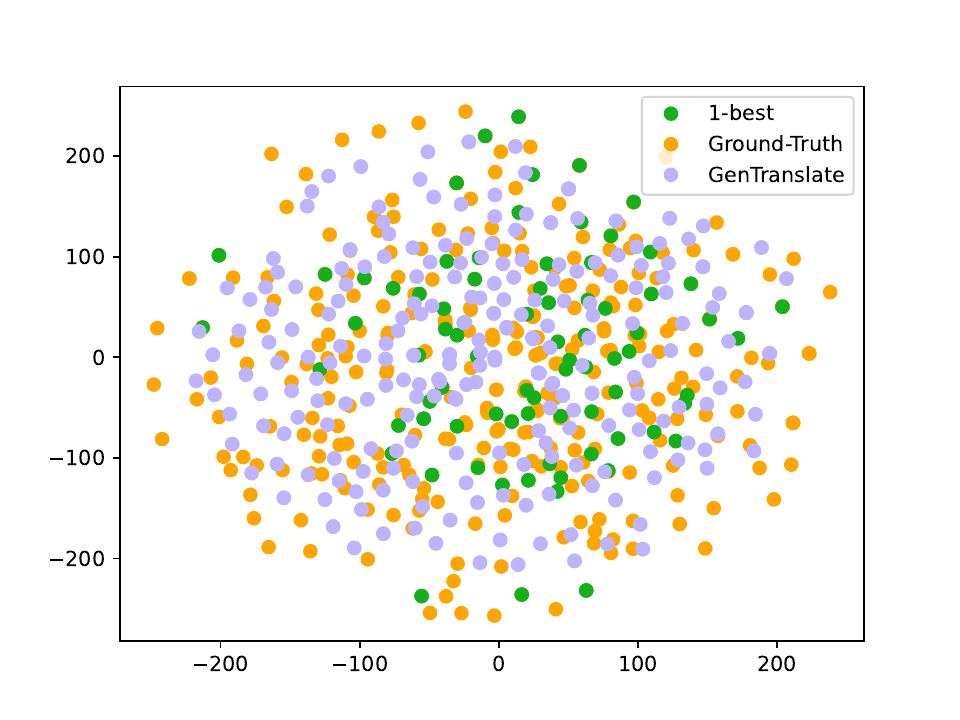}
\end{center}
\vspace{-0.3cm}
\caption{t-SNE visualization of n-grams in 1-best hypothesis ({\color{green}green}), ground-truth translation ({\color{orange}orange}) and GenTranslate output ({\color{violet}purple}).
It's an extension of Fig.~\ref{fig2}.}
\vspace{-0.2cm}
\label{fig5}
\end{figure}

Fig.~\ref{fig5} visualizes the n-gram tokens in GenTranslate output, which contains sufficient semantic information to match the ground-truth translation.
In comparison, the 1-best hypothesis lacks such information to produce high-quality translation output, which highlights the contribution of $N$-best hypotheses in GenTranslate paradigm (see Fig.~\ref{fig2}).

\section{Conclusion}
\label{sec:conclusion}
In this paper, we propose a generative paradigm for translation tasks, namely GenTranslate, which leverages LLMs to integrate the diverse candidates in the decoded $N$-best list and generate a higher-quality translation result.
Furthermore, we release a HypoTranslate dataset to support LLM finetuning, which contains over $592$K hypotheses-translation pairs in 11 languages.
Experimental evidence on various speech and machine translation benchmarks shows that our GenTranslate significantly outperforms the state-of-the-art model.

\section*{Limitations}
There are two limitations existed in this work.
First, the contribution of LLMs in our GenTranslate paradigm focuses on $N$-best hypotheses integration, while the translation part is actually done by SeamlessM4T model.
Experiment results in Table~\ref{table:asr_gentrans} also indicate that LLMs are good at $N$-best hypotheses integration and SeamlessM4T is good at translation.
Therefore, our future work could focus on how to better engage LLMs into the translation part to further improve the translation quality.
Another limitation is about the latest second version of SeamlessM4T released by Meta, which indicates a stronger baseline for GenTranslate.
In fact, our experiments had already been done on SeamlessM4T-Large before Meta released the latest SeamlessM4T-Large-V2 on November 30th, 2023.
For comprehensive evaluation, we also rerun our main experiments on this latest V2 backbone, and our GenTranslate has shown similar effectiveness on it (highlighted in gray in Table~\ref{table:st_xen_fleurs} to~\ref{table:mt_enx_wmt}).
For brevity, we prefer to leave the ablation study and analyses on SeamlessM4T-Large backbone only, as our GenTranslate paradigm has shown similar effectiveness and patterns on V1 and V2 backbones.

\section*{Ethics Statement}
This work does not pose any ethical issues.
All the data used in this paper are publicly available and are used under following licenses: Creative Commons BY 4.0 License, Creative Commons CC0 License, Creative Commons BY-NC-ND 4.0 License, and Creative Commons BY-SA 4.0 License.

\bibliography{anthology,custom}

\appendix

\clearpage

\section{HypoTranslate Dataset Details}
\label{asec:hypotrans_details}
In this section, we introduce the details of our proposed HypoTranslate dataset.
We first introduce the speech and machine translation corpora that we utilize to build HypoTranslate in \S\ref{assec:st_corpus} and \S\ref{assec:mt_corpus}.
Then, we present the dataset statistics in \S\ref{assec:statistics}.

\subsection{Speech Translation Corpus Selection}
\label{assec:st_corpus}
For speech translation task, we select three popular and public datasets that cover multiple languages:

\vspace{0.1cm}
\noindent\textbf{FLEURS}\footnote{\url{https://huggingface.co/datasets/google/fleurs}}~\citep{conneau2023fleurs}:
Few-shot Learning Evaluation of Universal Representations of Speech (FLEURS) benchmark provides an n-way parallel speech dataset in 102 languages built on top of the machine translation FLORES-101 benchmark~\citep{goyal2022flores}, with approximately 12 hours of speech supervision per language.
In this work, we select 15 X$\rightarrow$En and 6 En$\rightarrow$X language directions of speech translation data for evaluation.

\vspace{0.1cm}
\noindent\textbf{CoVoST-2}\footnote{\url{https://github.com/facebookresearch/covost}}~\citep{wang2020covost}:
CoVoST-2 is a popular multilingual speech translation corpus based on Common Voice~\citep{ardila2019common} that consists of 2,880 hours speech data recorded from 78K speakers. 
In this work, we select 15 X$\rightarrow$En and 3 En$\rightarrow$X language directions for evaluation.
Specifically, for En$\rightarrow$X language directions, we randomly select 1,000 testing samples from the original test split for higher evaluation efficiency.

\vspace{0.1cm}
\noindent\textbf{MuST-C}\footnote{\url{https://mt.fbk.eu/must-c-releases/}}~\citep{di2019must}:
MuST-C is a multilingual speech translation corpus whose size and quality facilitate the training of end-to-end systems for  spoken language translation from English into 15 languages.
In this work, we select 3 En$\rightarrow$X language directions for evaluation.

\subsection{Machine Translation Corpus Selection}
\label{assec:mt_corpus}
For machine translation task, we select two popular and public datasets that cover multiple languages:

\vspace{0.1cm}
\noindent\textbf{FLORES}\footnote{\url{https://huggingface.co/datasets/facebook/flores}}~\citep{costa2022no}:
FLORES consists of 3001 sentences sampled from English-language Wikimedia projects for 204 total languages. Approximately one third of sentences are collected from each of these sources: Wikinews, Wikijunior, and Wikivoyage. The content is professionally translated into 200+ languages to create FLORES dataset.
In this work, we select 10 X$\rightarrow$En language directions for evaluation.

\vspace{0.1cm}
\noindent\textbf{WMT}:
The Conference on Machine Translation (WMT) is a popular evaluation benchmark for MT task.
In this work, we select the newstest data of Ro$\rightarrow$En language direction from WMT'16\footnote{\url{https://www.statmt.org/wmt16/translation-task.html}}~\citep{bojar2016findings}, Cs$\rightarrow$En and It$\rightarrow$En directions from WMT'19\footnote{\url{https://www.statmt.org/wmt19/translation-task.html}}~\citep{barrault2019findings}, Ja$\rightarrow$En and Zh$\rightarrow$En directions from WMT'20\footnote{\url{https://www.statmt.org/wmt20/translation-task.html}}~\citep{loic2020findings} for evaluation, and corresponding newdev data is used for validation.
The training data is obtained from ParaCrawl-V9\footnote{\url{https://paracrawl.eu/}}~\citep{banon2020paracrawl} and JParaCrawl\footnote{\url{https://www.kecl.ntt.co.jp/icl/lirg/jparacrawl/}}~\citep{morishita2020jparacrawl} datasets.

\subsection{Statistics}
\label{assec:statistics}
After performing beam search decoding on the selected speech and machine translation corpora introduced above, we collect over 592K pairs of $N$-best hypotheses and ground-truth translation to build the HypoTranslate dataset. 
The statistics are illustrated in Table~\ref{table:statistics_st} and \ref{table:statistics_mt}, which present the number of hypotheses-translation pairs and the average utterance length. 
We plan to release the HypoTranslate dataset to public upon publication and open the development venue for more data.

\section{Experimental Setup Details}
\label{asec:exp_setup}

\begin{table*}[t]
\centering
\resizebox{0.88\textwidth}{!}{
\begin{tabular}{c|c|c|c|c}
\toprule[1.2pt]
LLM & LLaMA-2-7b & LLaMA-2-13b & BigTranslate & ALMA-13b \\
\midrule[1.2pt]
Number of Transformer Layers $H$ & 32 & 40 & 40 & 40 \\
Number of Attention Heads $N_\text{head}$ & 32 & 40 & 40 & 40 \\
Embedding Size $D$ & 4,096 & 5,120 & 5,120 & 5,120 \\
Block Size $B$ & 4,096 & 4,096 & 4,096 & 4,096 \\
Vocabulary Size $V$ & 32,000 & 32,000 & 53,613 & 32,000 \\
\bottomrule[1.2pt]
\end{tabular}}
\caption{Comparison between main configurations of different popular LLMs.}
\label{table:llm_config}
\end{table*}

\subsection{Model Setups}
\label{assec:model_setups}
We select two latest foundation LLMs for evaluation, including LLaMA-2-7b~\citep{touvron2023llama2} and LLaMA-2-13b~\citep{touvron2023llama2}.
In addition, in order to evaluate the multilingual ability of LLMs for GenTranslate with non-English-target directions, we also select two latest finetuned LLMs on MT task, including BigTranslate~\citep{yang2023bigtrans} and ALMA-13b~\citep{xu2023paradigm}.
Table~\ref{table:llm_config} compares their main configurations.
For efficient LLM finetuning, we follow the default settings of LLaMA-Adapter\footnote{\url{https://github.com/Lightning-AI/lit-gpt/blob/main/lit_gpt/adapter.py}}~\citep{zhang2023llama}.

\subsection{Inference Setups}
\label{assec:infer}
In the response generation during inference stage, we set a temperature of 0.2 and top-1 sampling, \emph{i.e.}, greedy search.
We have observed over-confidence phenomenon in our experiments (\emph{i.e.}, output probability distribution for decision is close to one-hot), which results in similar performance with different $k$ for top-$k$ sampling.
Therefore, we select top-1 sampling for higher decoding efficiency.




\subsection{Translation Baselines}
\label{assec:baselines}
To comprehensively evaluate our GenTranslate model, we selected some of the latest and most advanced baselines in speech and machine translation for comparison. We will introduce these in the following subsections.

\subsubsection{Speech Translation}

\vspace{0.1cm}
\noindent\textbf{XLS-R}\footnote{\url{https://huggingface.co/models?other=xls_r}}~\citep{babu2021xls}:
XLS-R is a large-scale model for cross-lingual speech representation learning based on Wav2vec 2.0~\citep{baevski2020wav2vec}.
They train models with up to 2B parameters on 500K hours of publicly available speech audio in 128 languages, which achieves superior performance on a wide range of multilingual speech processing tasks, including speech translation, speech recognition and language identification.

\vspace{0.1cm}
\noindent\textbf{Whisper}\footnote{\url{https://github.com/openai/whisper}}~\citep{radford2023robust}:
Whisper is a large-scale automatic speech recognition (ASR) system~\cite{chen2022noise,chen2022self,chen2023leveraging,chen2023metric,hu2022interactive,hu2022dual,hu2023gradient,hu2023wav2code,hu2023hearing,hu2023mir,hu2023cross,hu2023noise,zhu2023robust,zhu2024multichannel} trained on 680K hours of multilingual and multitask supervised data collected from the web, which shows excellent robustness to accents, background noise and technical language. Moreover, it enables transcription in multiple languages, as well as translation from those languages into English.

\vspace{0.1cm}
\noindent\textbf{AudioPaLM2}~\citep{rubenstein2023audiopalm}: AudioPaLM2 fuses text-based and speech-based language models, PaLM-2~\citep{anil2023palm} and AudioLM~\citep{borsos2023audiolm}, into a unified multimodal architecture that can process and generate text and speech with applications including speech recognition and speech-to-speech translation. 
AudioPaLM2 inherits the capability to preserve paralinguistic information such as speaker identity and intonation from AudioLM and the linguistic knowledge present only in text large language models such as PaLM-2.
The resulting model significantly outperforms existing systems for speech translation and has the ability to perform zero-shot speech-to-text translation for many unseen languages.

\noindent\textbf{ComSL}\footnote{\url{https://github.com/nethermanpro/ComSL}}~\citep{le2023comsl}:
ComSL is a speech-language model built atop a composite architecture of public pre-trained speech-only and language-only models and optimized data-efficiently for spoken language tasks.
Particularly, they propose to incorporate cross-modality learning into transfer learning and conduct them simultaneously for downstream tasks in a multi-task learning manner, which has demonstrated effectiveness in end-to-end speech-to-text translation tasks.

\subsubsection{Machine Translation}

\noindent\textbf{NLLB}\footnote{\url{https://huggingface.co/facebook/nllb-200-3.3B}}~\citep{costa2022no}:
No Language Left Behind (NLLB) is a first-of-its-kind, AI breakthrough project that open-sources models capable of delivering evaluated, high-quality translations directly between 200 languages.

\noindent\textbf{BigTranslate}\textsuperscript{\ref{fn5}}~\citep{yang2023bigtrans}:
BigTranslate adapts LLaMA-13b~\citep{touvron2023llama} that covers only 20 languages and enhances it with multilingual translation capability on up to 102 languages by instruction-following finetuning, which achieves comparable MT performance to ChatGPT~\citep{chatgpt} and Google Translate.

\noindent\textbf{ALMA}\textsuperscript{\ref{fn6}}~\citep{xu2023paradigm}:
ALMA proposes a novel finetuning approach for LLMs that is specifically designed for MT task, eliminating the need for the abundant parallel data that traditional translation models usually depend on, which includes two stages: initial finetuning on monolingual data followed by subsequent finetuning on a small set of high-quality parallel data.
Built based on LLaMA-2, it has achieved significant improvement over prior works across multiple translation directions.

\section{Supplementary Experiment Results}
\label{asec:supple_exp}


\begin{table*}[t]
\centering
\resizebox{0.93\textwidth}{!}{
\begin{tabular}{c|p{9cm}|c}
\toprule[1.2pt]
Method & Utterance & BLEU Score \\
\midrule[1.2pt]
\multirow{5}{*}{$N$-best Candidates} & 地球河流流入海洋的20\%的水来自亚马逊. & 15.0 \\
& 地球河流流入海洋的20\%的水源来自亚马逊. & 15.0 \\
& 地球河流流入海洋的全部20\%的水来自亚马逊. & 12.3 \\
& 地球河流流入海洋的20\%的水来自亚马逊 & 15.0 \\
& 地球河流流入海洋的全部20\%的水来自亚马逊 & 12.3 \\
\midrule
GenTranslate (ours) & 地球上的河流汇入大洋的 20\% 的水来自亚马逊河。 & \textbf{18.7} \\
\midrule
Ground-truth Translation & 亚马逊河占全世界所有河流的入海流量的 20\%。 & - \\
\bottomrule[1.2pt]
\end{tabular}}
\caption{Supplementary case study.
The test sample is selected from the FLEURS En$\rightarrow$Zh ST test set.
}
\label{table:supply_case_study}
\end{table*}

\begin{table}[t]
\centering
\resizebox{1.0\columnwidth}{!}{
\begin{tabular}{p{3cm}cc}
\toprule[1.2pt]
Language & Family & Sub-grouping \\
\midrule[1.2pt]
Persian (Fa) & Indo-European & Indo-Iranian \\
Hindi (Hi) & Indo-European & Indo-Iranian \\
Italian (It) & Indo-European & Indo-Iranian \\
Spanish (Es) & Indo-European & Italic \\
French (Fr) & Indo-European & Italic \\
Portuguese (Pt) & Indo-European & Italic \\
Welsh (Cy) & Indo-European & Celtic \\
English (En) & Indo-European & Germantic \\
German (De) & Indo-European & Germantic \\
Greek (El) & Indo-European & Greek \\
Ukranian (Uk) & Indo-European & Balto-Slavic \\
Arabic (Ar) & Afro-Asiatic & Semitic \\
Vietnamese (Vi) & Austro-Asiatic & Mon-Khmer \\
Japanese (Ja) & Japonic & - \\
Tamil (Ta) & Dravidian & Dravidian \\
Chinese (Zh) & Sino-Tibetan & Chinese \\
\bottomrule[1.2pt]
\end{tabular}}
\vspace{-0.1cm}
\caption{Detailed language family and sub-grouping information~\citep{babu2021xls} of FLEURS datasets.
}
\label{table:detailed_family_info}
\vspace{-0.3cm}
\end{table}

\subsection{Supplementary Case Study}
\label{assec:supply_case_study}
Table~\ref{table:supply_case_study} supplies a case study from FLEURS En$\rightarrow$Zh ST test set.
We can observe that the $N$-best candidate are semantically similar to each other and only varies in sentence structure.
In our GenTranslate paradigm, LLMs integrates the different patterns of $N$-best hypotheses to generate a new translation result with 3.7 BLEU improvement over 1-best hypothesis.
Such observation verifies the effectiveness of LLMs in our GenTranslate paradigm to generate better translation output.

\subsection{BLEU vs. chrF++}
\label{assec:chrf}
We report translation performance in terms of the BLEU score~\citep{papineni2002bleu} in most experiments of this work.
For more comprehensive evaluation, Table~\ref{table:st_xen_fleurs_chrf} presents both BLEU and chrF++ scores~\citep{popovic2017chrf++,barrault2023seamlessm4t} on FLEURS X$\rightarrow$En test sets, where we can observe consistent improvements of BLEU and chrF++ scores (2.1 $\Delta$ BLEU and 0.9 $\Delta$ chrF++) in GenTranslate.
It indicates that both metrics are applicable for the evaluation of translation tasks.

\begin{figure}[t!]
\begin{center}
\includegraphics[width=0.97\columnwidth]{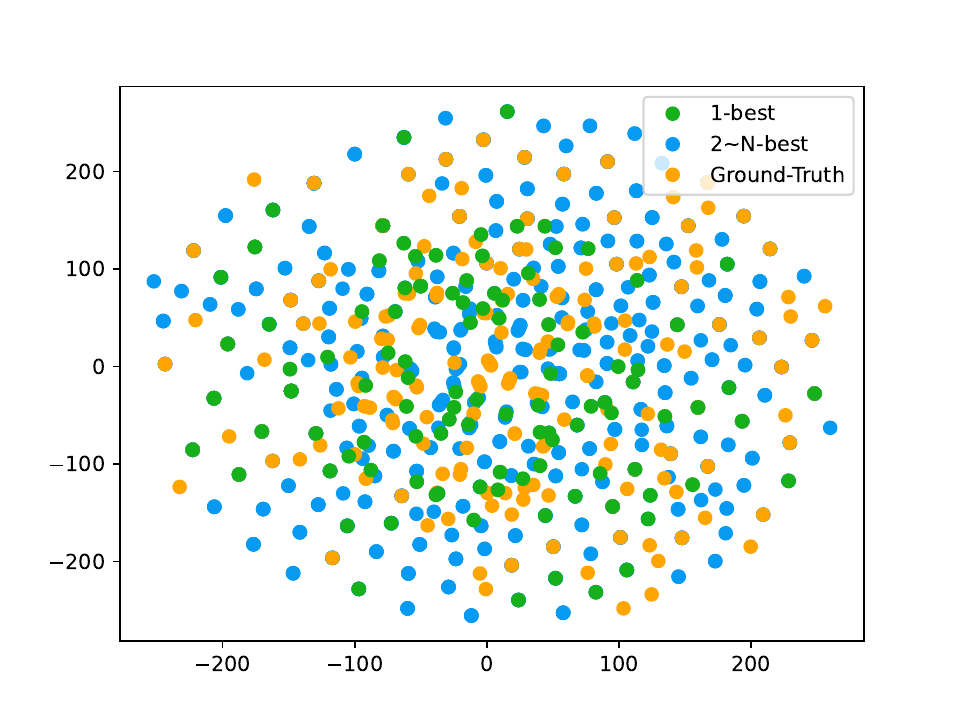}
\end{center}
\vspace{-0.2cm}
\caption{t-SNE visualization of n-gram tokens in ASR 1-best hypothesis ({\color{green}green}), 2 to $N$-best hypotheses ({\color{blue}blue}), and the ground-truth transcription ({\color{orange}orange}).
Different from the ST hypotheses in Fig.~\ref{fig2}, ASR 1-best hypothesis aligns well with the ground-truth transcription, where the role of 2$\sim$$N$-best hypotheses is to provide diverse candidate tokens for correcting errors.
}
\vspace{-0.3cm}
\label{fig6}
\end{figure}

\begin{table*}[t]
\centering
\resizebox{1.0\textwidth}{!}{
\begin{tabular}{cp{2.8cm}|cc|cc|cc}
\toprule[1.2pt]
\multirow{2}{*}{Data Source} & Source / Target & \multicolumn{2}{c|}{Train} & \multicolumn{2}{c|}{Dev.} & \multicolumn{2}{c}{Test} \\
& Language X & \# Pairs & Length & \# Pairs & Length & \# Pairs & Length \\
\midrule[1.2pt]
\multirow{15}{*}{\begin{tabular}[c]{@{}c@{}}FLEURS~\cite{conneau2023fleurs} \\ (X$\rightarrow$En)\end{tabular}} & Arabic (Ar) & 2,062 & 20.4 & 295 & 19.8 & 428 & 21.4 \\ 
& Welsh (Cy) & 3,349 & 21.1 & 447 & 20.6 & 1,021 & 22.1 \\ 
& German (De) & 2,926 & 20.7 & 363 & 20.1 & 862 & 21.9 \\ 
& Greek (El) & 3,148 & 20.9 & 271 & 20.5 & 650 & 21.7 \\ 
& Spanish (Es) & 2,732 & 20.8 & 408 & 20.5 & 908 & 21.8 \\ 
& Persian (Fa) & 3,032 & 20.9 & 369 & 20.1 & 871 & 21.8 \\ 
& French (Fr) & 3,119 & 20.8 & 289 & 19.9 & 676 & 21.8 \\ 
& Hindi (Hi) & 2,072 & 20.6 & 239 & 19.2 & 418 & 21.4 \\ 
& Italian (It) & 2,970 & 20.6 & 391 & 20.2 & 865 & 21.7 \\ 
& Japanese (Ja) & 2,241 & 20.2 & 266 & 19.6 & 650 & 21.3 \\ 
& Portuguese (Pt) & 2,731 & 20.7 & 386 & 20.2 & 919 & 21.9 \\ 
& Tamil (Ta) & 2,317 & 20.7 & 377 & 20.0 & 591 & 22.0 \\ 
& Ukrainian (Uk) & 2,741 & 20.8 & 325 & 20.3 & 750 & 22.0 \\ 
& Vietnamese (Vi) & 2,927 & 20.7 & 361 & 20.2 & 857 & 21.8 \\ 
& Chinese (Zh) & 3,178 & 21.0 & 409 & 20.6 & 945 & 22.1 \\ 
\midrule
\multirow{15}{*}{\begin{tabular}[c]{@{}c@{}}CoVoST-2~\cite{wang2020covost} \\ (X$\rightarrow$En)\end{tabular}} & French (Fr) & 30,000 & 8.9 & 1,000 & 8.9 & 14,760 & 9.4 \\ 
& German (De) & 30,000 & 9.8 & 1,000 & 10.2 & 13,511 & 9.8 \\ 
& Catalan (Ca) & 30,000 & 10.3 & 1,000 & 10.3 & 12,730 & 10.5 \\ 
& Spanish (Es) & 30,000 & 9.7 & 1,000 & 9.6 & 13,221 & 9.9 \\ 
& Russian (Ru) & 12,112 & 11.9 & 1,000 & 11.9 & 6,300 & 11.8 \\ 
& Chinese (Zh) & 7,085 & 12.0 & 1,000 & 11.9 & 4,898 & 11.6 \\ 
& Dutch (Nl) & 7,108 & 8.2 & 1,000 & 8.5 & 1,699 & 8.5 \\ 
& Turkish (Tr) & 3,966 & 8.3 & 1,000 & 8.1 & 1,629 & 8.3 \\ 
& Estonian (Et) & 1,782 & 17.8 & 1,000 & 15.5 & 1,571 & 16.1 \\ 
& Mongolian (Mn) & 2,067 & 11.2 & 1,000 & 11.2 & 1,759 & 11.3 \\ 
& Arabic (Ar) & 2,283 & 5.8 & 1,000 & 5.7 & 1,695 & 5.7 \\ 
& Latvian (Lv) & 2,337 & 6.1 & 1,000 & 6.3 & 1,629 & 6.2 \\ 
& Slovenian (Sl) & 1,843 & 7.2 & 509 & 7.0 & 360 & 6.3 \\ 
& Japanese (Ja) & 1,119 & 8.3 & 635 & 8.5 & 684 & 8.4 \\ 
& Indonesian (Id) & 1,243 & 6.6 & 792 & 6.6 & 844 & 6.7 \\ 
\midrule
\multirow{6}{*}{\begin{tabular}[c]{@{}c@{}}FLEURS~\cite{conneau2023fleurs} \\ (En$\rightarrow$X)\end{tabular}} & Spanish (Es) & 2,502 & 25.0 & 394 & 25.1 & 643 & 26.1 \\
& French (Fr) & 2,592 & 24.4 & 363 & 24.1 & 612 & 25.5 \\
& Italian (It) & 2,564 & 23.2 & 386 & 22.8 & 640 & 24.4 \\
& Japanese (Ja) & 2,290 & 53.6 & 351 & 53.1 & 592 & 55.6 \\
& Portuguese (Pt) & 2,503 & 22.4 & 387 & 21.9 & 645 & 23.4 \\
& Chinese (Zh) & 2,592 & 42.3 & 394 & 40.7 & 646 & 42.7 \\
\midrule
\multirow{3}{*}{\begin{tabular}[c]{@{}c@{}}CoVoST-2~\cite{wang2020covost} \\ (En$\rightarrow$X)\end{tabular}} & Persian (Fa) & 30,000 & 10.8 & 1,000 & 9.3 & 1,000 & 9.5 \\
& Japanese (Ja) & 30,000 & 28.5 & 1,000 & 26.6 & 1,000 & 23.3 \\
& Chinese (Zh) & 30,000 & 19.7 & 1,000 & 19.7 & 1,000 & 16.0 \\
\midrule
\multirow{3}{*}{\begin{tabular}[c]{@{}c@{}}MuST-C~\cite{di2019must} \\ (En$\rightarrow$X)\end{tabular}} & Spanish (Es) & 6,000 & 19.4 & 1,316 & 20.1 & 2,502 & 17.1 \\ 
& Italian (It) & 6,000 & 18.2 & 1,309 & 18.8 & 2,574 & 16.4 \\
& Chinese (Zh) & 6,000 & 49.6 & 888 & 63.7 & 1,823 & 46.3 \\
\midrule
\multicolumn{2}{c|}{Total} & 327,533 & 15.9 & 27,920 & 16.9 & 102,378 & 13.3 \\
\bottomrule[1.2pt]
\end{tabular}}
\caption{HypoTranslate dataset \textbf{(ST part)} statistics in terms of the number of hypotheses-translation pairs and average length of ground-truth utterance in different language directions.
}
\label{table:statistics_st}
\end{table*}

\begin{table*}[t!]
\centering
\resizebox{1.0\textwidth}{!}{
\begin{tabular}{l|ccccccccccccccc|c}
\toprule[1.2pt]
X$\rightarrow$En & Ar & Cy & De & El & Es & Fa & Fr & Hi & It & Ja & Pt & Ta & Uk & Vi & Zh & Avg. \\
\midrule[1.2pt]
\multicolumn{17}{l}{\emph{\textbf{BLEU score}}} \\
SeamlessM4T (ASR+MT) & 38.9 & 37.0 & 39.7 & 29.0 & 27.7 & 34.1 & 37.7 & 33.9 & 28.9 & 21.7 & 42.3 & 23.7 & 34.0 & 24.9 & 24.4 & 31.9 \\
GenTranslate (ours) & {\textbf{39.9}} & {\textbf{39.4}} & {\textbf{41.6}} & {\textbf{32.8}} & {\textbf{31.2}} & {\textbf{35.9}} & {\textbf{40.6}} & {\textbf{34.9}} & {\textbf{32.1}} & \textbf{22.8} & {\textbf{45.0}} & {\textbf{24.1}} & {\textbf{36.9}} & {\textbf{27.4}} & {\textbf{25.7}} & {\textbf{34.0}} \\
\midrule
\multicolumn{17}{l}{\emph{\textbf{chrF++ score}}} \\
SeamlessM4T (ASR+MT) & 62.7 & 60.0 & 63.8 & 55.0 & 56.0 & 58.7 & 62.4 & 58.8 & 57.0 & \textbf{47.9} & 65.9 & \textbf{49.8} & 59.2 & 50.5 & 51.5 & 57.3 \\
GenTranslate (ours) & {\textbf{63.1}} & {\textbf{61.2}} & {\textbf{64.9}} & {\textbf{57.0}} & {\textbf{57.1}} & {\textbf{59.7}} & {\textbf{64.0}} & {\textbf{59.1}} & {\textbf{58.0}} & 47.6 & {\textbf{67.2}} & 49.7 & {\textbf{60.8}} & {\textbf{51.6}} & {\textbf{52.0}} & {\textbf{58.2}} \\
\bottomrule[1.2pt]
\end{tabular}}
\caption{Speech translation results on FLEURS X$\rightarrow$En test sets in terms of chrF++ score.
}
\label{table:st_xen_fleurs_chrf}
\vspace{-0.1cm}
\end{table*}

\begin{table*}[t]
\centering
\resizebox{1.0\textwidth}{!}{
\begin{tabular}{cp{2.6cm}|cc|cc|cc}
\toprule[1.2pt]
\multirow{2}{*}{Data Source} & Source / Target & \multicolumn{2}{c|}{Train} & \multicolumn{2}{c|}{Dev.} & \multicolumn{2}{c}{Test} \\
& Language X & \# Pairs & Length & \# Pairs & Length & \# Pairs & Length \\
\midrule[1.2pt]
\multirow{10}{*}{\begin{tabular}[c]{@{}c@{}}FLORES~\cite{costa2022no} \\ (X$\rightarrow$En)\end{tabular}} & Arabic (Ar) & 2,062 & 20.4 & 295 & 19.8 & 1,012 & 21.6 \\ 
& German (De) & 2,926 & 20.7 & 363 & 20.1 & 1,012 & 21.6 \\ 
& Greek (El) & 3,148 & 20.9 & 271 & 20.5 & 1,012 & 21.6 \\ 
& Spanish (Es) & 2,732 & 20.8 & 408 & 20.5 & 1,012 & 21.6 \\ 
& Persian (Fa) & 3,032 & 20.9 & 369 & 20.1 & 1,012 & 21.6 \\ 
& French (Fr) & 3,119 & 20.8 & 289 & 19.9 & 1,012 & 21.6 \\ 
& Italian (It) & 2,970 & 20.6 & 391 & 20.2 & 1,012 & 21.6 \\ 
& Japanese (Ja) & 2,241 & 20.2 & 266 & 19.6 & 1,012 & 21.6 \\  
& Ukrainian (Uk) & 2,741 & 20.8 & 325 & 20.3 & 1,012 & 21.6 \\ 
& Chinese (Zh) & 3,178 & 21.0 & 409 & 20.6 & 1,012 & 21.6 \\ 
\midrule
\multirow{5}{*}{\begin{tabular}[c]{@{}c@{}}WMT'\{16,19,20\} \\ (En$\rightarrow$X)\end{tabular}} & Czech (Cs) & 15,000 & 12.3 & 2,983 & 15.8 & 1,997 & 18.8 \\
& Japanese (Ja) & 15,000 & 49.8 & 1,998 & 53.1 & 1,000 & 59.8 \\
& Lithuanian (Lt) & 15,000 & 12.0 & 2,000 & 16.5 & 998 & 16.6 \\
& Romanian (Ro) & 15,000 & 16.7 & 1,999 & 22.6 & 1,999 & 21.7 \\
& Chinese (Zh) & 15,000 & 35.6 & 1,997 & 47.8 & 1,418 & 60.7 \\
\midrule
\multicolumn{2}{c|}{Total} & 103,149 & 24.0 & 14,363 & 27.5 & 17,532 & 26.3 \\
\bottomrule[1.2pt]
\end{tabular}}
\caption{HypoTranslate dataset \textbf{(MT part)} statistics in terms of the number of hypotheses-translation pairs and average length of ground-truth utterance in different language directions.
}
\label{table:statistics_mt}
\end{table*}

\end{CJK}
\end{document}